\documentclass{article}

\usepackage[preprint]{neurips_2026}

\usepackage[utf8]{inputenc} % allow utf-8 input
\usepackage[T1]{fontenc}    % use 8-bit T1 fonts
\usepackage{hyperref}       % hyperlinks
\usepackage{url}            % simple URL typesetting
\usepackage{booktabs}       % professional-quality tables
\usepackage{amsfonts}       % blackboard math symbols
\usepackage{nicefrac}       % compact symbols for 1/2, etc.
\usepackage{microtype}      % microtypography
\usepackage{xcolor}         % colors
\usepackage{natbib}
\usepackage{graphicx}
\usepackage{caption}
\usepackage{subcaption}
\usepackage{array}
\usepackage{wrapfig}
\usepackage{placeins}
\setcitestyle{numbers,square}

\title{Low-Frequency Shortcuts in Texture-Driven \\ Visual Learning}

\author{
  Utku \c{S}irin \\
  Harvard University
  \And
  Cathy Hou \\
  Harvard University \\
  \And
  Despina-Ekaterini Argiropoulos\thanks{Work done during an internship at Harvard University.} \\
  Foundation for Research and Technology-Hellas \\
  University of Crete \\
  \And
  David Alvarez-Melis \\
  Harvard University \\
  Kempner Institute \\
  \And
  Stratos Idreos \\
  Harvard University \\
}

\begin{document}

\maketitle

\begin{abstract}

Neural networks suffer from shortcut learning, where learned features generalize well to the training set but not to in-distribution (ID) or out-of-distribution (OOD) test sets. Existing studies are all based on a few standard benchmarks, which are shape-driven. Numerous application domains, however, are texture-driven. In this work, we present shortcut learning analysis for texture-driven domains, and compare it with that of a standard benchmark. We show that texture-driven domains suffer from low-frequency shortcuts. They make the majority of their decisions based on a few low-frequency components (LFCs) with a skewed spectral behavior, despite that higher-frequency components (HFCs) have a higher predictive power. Pruning LFCs from training and test sets mitigates the shortcut and provides a more balanced spectral behavior, improving the ID accuracy by up to 10\% and OOD accuracy by up to 40\% under algorithmic and real-world domain shifts.
We show that general-purpose and domain-specific foundation models can also suffer from low-frequency shortcuts. While large models can mitigate the shortcuts, they incur a high computational cost and may result in a significantly lower accuracy than shortcut-pruned from-scratch trained small models.
We show that reduced image resolutions amplify the degree of shortcuts; large frequency-transformation block sizes capture low-frequency shortcuts better than small block sizes; and, low-frequency shortcuts persist across different color spaces. Our findings provide valuable insights, which we hope will be useful for practitioners working on new, understudied domains.

%We show that low-frequency shortcuts make the models highly vulnerable to OOD corruptions, leading up to 70\% accuracy drop compared to the ID accuracy. Pruning LFCs significantly improves robustness to low-frequency corruptions, by up to 40\%, and introduces a trade-off for high-frequency corruptions; the balanced spectral behavior provides a better generalization performance, whereas the increased dependence on high-frequency features reduces it.  OOD accuracy depends on the interaction between these two factors. 

 %Low-frequency shortcuts are pathological spectral behaviors, where accuracy-contributions of individual frequency components are highly skewed towards low frequencies. 
  
  % Low-frequency shortcuts persist across different model sizes, architectures, and hyperparameters. 

  % Using heavy training sets robust features cannot avoid. Are not enough. Even pretrained VFM/VLMs suffer from low-frequency shortcuts
    
  %(1) Shape-driven tasks, in contrast, benefit from LFCs as LFCs capture smooth features that underlie shape structures. (2) We show that low-frequency shortcuts are independent of the used model architecture, model size, and optimizer. (3) Even pre-trained vision foundation and vision-language models can suffer from low-frequency shortcuts, despite their heavily tuned model weights across various data augmentations and training sets. (4) Even though ID decreases, OOD heavily increases thanks to eliminating the skew. Skew is the problem. (5) Mixed semantics contribution.
\end{abstract}

\section{Introduction}
\label{sect:intro}

\noindent \textbf{Simplicity bias \& shortcut learning.} Neural networks are biased towards learning simple functions that solve the optimization problem fast, rather than complex functions that are faithful to the semantics of the problem \cite{dagaev2023, geirhos2020, nauta2022, pezeshki2021, rahaman2019, shah2020, shortcut1, shortcut2, zhou2021}. This results in shortcut learning, where learned features are strong predictors in the validation set, but do not generalize well into out-of-distribution (OOD) test sets.
%\footnote{ID generalization problems are commonly referred to as overfitting, whereas OOD generalization problems are referred to as shortcutting \cite{geirhos2020}. In this work, we refer to both phenomena as shortcutting, as we are interested in features in the training set that are easy to learn, but do not generalize well to ID or OOD test sets. Simplicity bias can hurt ID generalization as well as OOD generalization, as also observed by \cite{shah2020} (see their Section 5), and by \cite{gavrikov2024} (see their Section 5.1).}. 
For example, neural networks might learn decision rules based on superficial cues, such as the source tag in horse images in the Pascal VOC dataset \cite{beery2018}, and hence suffer from low generalization performance and unreliable predictions.

\noindent \textbf{Frequency analysis for shortcut learning.} Simplicity bias might occur due to visible \cite{beery2018,boland2024} or invisible superficial cues \cite{shortcut1,shortcut2}. While visible features are easy to detect via visual inspection, invisible features are deeply embedded in the data and require formal tools to analyze \cite{bau2017, dabounou2024, papernot2018, ramaswamy2023, tuo2025, shortcut1, shortcut2, zhao2019}. One such useful tool is Fourier analysis \cite{jo2017,wang2020,shortcut1}. Fourier analysis transforms the data from its original domain, e.g., the spatial domain for images, into the frequency domain. Frequency representation provides a structured view of the data and enables principled analysis of learning dynamics by leveraging the data's frequency structure. 

% %We show that texture-driven domains make majority of their decisions based on a few low-frequency components (LFCs), despite that their classification information is in higher-frequency features with fine-grained, repetitive patterns. We call this phenomenon as low-frequency shortcuts. Pruning LFCs mitigates the shortcut and provide a more balanced spectral behavior, which in turn provides up to 8\% higher ID accuracy. We show that skewed spectral behavior further makes the models highly vulnerable to OOD corruptions, causing up to 70\% reduced accuracy compared to ID performance. Pruning LFCs significantly improves the robustness to low-frequency corruptions, and introduces a trade-off for high-frequency corruptions; the balanced spectral behavior improves the generalization performance, whereas the increased dependence on the higher-frequency ranges decreases it. OOD accuracy depends on the interaction between these two factors.
\begin{figure*}[t]
  \centering
   \includegraphics[width=1\linewidth]{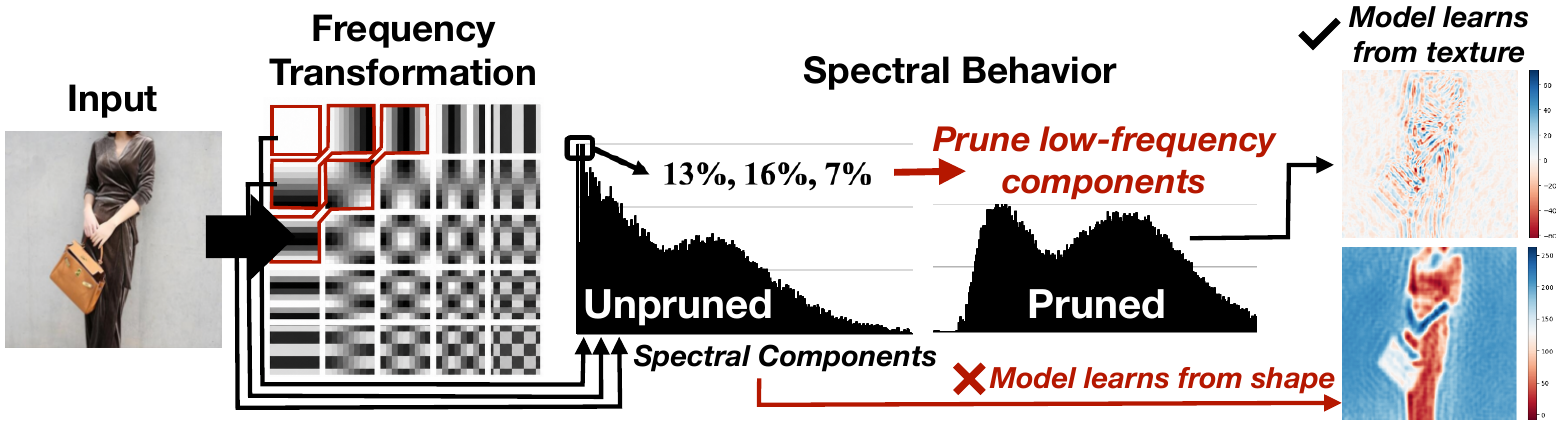}
   \caption{
    Spectral behavior of models trained on texture-driven domains show that models make majority of their decisions based on a few low-frequency components (first three figures above, from left). Once we prune low-frequency components (LFCs) from training and test sets, spectral behavior shifts towards higher-frequencies (forth figure above), and we obtain a significantly higher ID accuracy, up to 10\% (Section \ref{sect:lfc_shortcuts}). These results show that texture-driven domains suffer from low-frequency shortcuts, where models learn from LFCs despite that higher-frequency components have a higher predictive power. We show that low-frequency shortcuts persist across different model sizes, architectures, and hyperparameters (Section \ref{sect:lfc_shortcuts}), under various OOD conditions (Section \ref{sect:ood} \& \ref{sect:mixed_semantics}), for pretrained foundation models (Section \ref{sect:vlm_vfm}), and different resolutions, transformation block sizes, and color spaces (Section \ref{sect:resolution_et_al}).
   }
   \label{fig:main_figure}
\end{figure*}

\noindent \textbf{Existing studies are limited to shape semantics.} Existing frequency analysis studies are mostly based on a few standard benchmarks, such as CIFAR-10 and ImageNet \cite{jo2017,wang2020,shortcut1}. Standard benchmarks consist of natural images of everyday objects, where classification information is centered around specific objects with well-known global shapes. Therefore, known shortcuts, such as spectral bias \cite{rahaman2019}, texture bias \cite{geirhos2019}, or shape sensitivity \cite{burgert2025}, are all relative to object-centric representations, which assume that shape is the dominant cue and analyze how frequency components interfere with or support the shape semantics. 

\noindent \textbf{Numerous application domains are texture-driven.} Computer vision applications, however, span a wide range of domains that are texture-driven, such as histopathology \cite{spider2025} and textile classification \cite{textilenet2023}.
These domains include images, such as microscopic tissue scans or textile samples, where the notion of ``shape'' is ill-defined or irrelevant (see Figure \ref{fig:example_domains}). Classification information is distributed across the image, with repetitive fine-grained patterns rather than being centered on well-defined objects. Spectral cues that drive model predictions shift from global shape structures to fine-grained spatial patterns. As a result, whether known or new shortcuts exist in texture-driven domains, and, if so, how they affect model predictions and robustness, is unclear.

%they do not contain well-defined objects, and their semantics arise from fine-grained spatial patterns rather than global shape structures. In these domains, the notion of ``shape'' is ill-defined or irrelevant

\noindent \textbf{Shortcut learning for texture-driven domains.} We study shortcut learning in texture-driven domains and compare it to a standard benchmark using frequency analysis tools. We consider histopathology, textile classification, ground terrain recognition, AI-generated images, morphological classification of galaxies, and land use/cover classification, analyzing one image classification task per domain. As a standard benchmark, we use CIFAR-10 \cite{cifar10} due to its widespread adoption in frequency-based studies \cite{abello2021,jo2017,wang2020}. 
We perform a comprehensive frequency-spectrum analysis covering low, mid, and high frequencies, using different model architectures, sizes, and hyperparameters, as well as pretrained, general-purpose and domain-specific vision foundation/language models, based on in-distribution (ID) and out-of-distribution (OOD) performance. In summary, our \textbf{contributions} are as follows.

\begin{itemize}

    \item We show that texture-driven domains suffer from low-frequency shortcuts. Neural networks make the majority of their decisions based on a few low-frequency components (LFCs), despite that higher-frequency components (HFCs) have higher predictive power. This results in severely suboptimal ID as well as OOD generalization performance. 

    \item We show that pruning LFCs can mitigate the shortcuts and provide up to 10\% higher ID and 40\% higher OOD performance across different model architectures, sizes, optimizers, and hyperparameters, under various OOD conditions. To the best of our knowledge, we are the first study showing that shortcuts can hurt ID performance, as well as OOD performance.\footnote{In their theoretical study, Shah et al. \cite{shah2020} also have shown that simplicity bias can hurt both ID and OOD performance, but using only a synthetic toy dataset and a single-layer simple neural network (see their Section 5 and Table 2). We show that simplicity bias is not only an issue for highly controlled toy scenarios, but actually a real-world phenomenon across a wide range of real-world datasets and models.}.
    
    %We show that low-frequency shortcuts result in severely suboptimal ID generalization performance. Neural networks learn from simple but less representative low-frequency features, instead of complex but more representative higher-frequency features. Removing the shortcuts from training and test sets allows neural networks to learn from higher-frequency features and hence achieve up to 10\% higher ID performance. 
    %"We are the first". Repetition with \#1?
    
    %We identify low-frequency shortcuts as a pathological spectral behavior, where accuracy-contributions of individual frequency components are highly skewed towards low frequencies. Pruning LFCs from training and test sets mitigates the shortcut and provide a more balanced spectral behavior over higher frequency features, which in turn provides up to 8\% higher ID accuracy.

    \item We study general-purpose and domain-specific VFM/VLMs and show that they too can suffer from low-frequency shortcuts. While large models can mitigate the shortcuts, they incur a high computational cost and may result in a significantly lower accuracy than shortcut-pruned from-scratch trained small models. 
    %Shortcut-pruned, from-scratch trained small models can provide a close-to-VFM accuracy at a fraction of the computational cost.
    
    %We show that low-frequency shortcuts make the models highly vulnerable to OOD test images due to their skewed reliance on low-frequency features, under both algorithmic and real-world distribution shifts. Removing the shortcuts from training and test sets allows models to learn from a wide-range of frequency components, making them significantly more robust to distribution shifts. 
    
    %We show that low-frequency shortcuts make the models highly vulnerable to OOD corruptions, leading up to 70\% accuracy drop compared to the ID accuracy. Pruning LFCs significantly improves robustness to low-frequency corruptions, by up to 40\%, and introduces a trade-off for high-frequency corruptions; the improved spectral behavior provides a better generalization performance, whereas the increased dependence on higher-frequency features reduces it. OOD accuracy depends on the interaction between these two factors.

    \item We study low-frequency shortcuts across different image resolutions, frequency-transformation block sizes, and color spaces. We show that reduced image resolutions increase the degree of shortcuts; small block sizes only partially mitigate the shortcuts; and, shortcuts persist across different color spaces.

\end{itemize}

%\vspace{-1.5em}

Our study provides valuable insights and improves our understanding of the learning dynamics of neural networks. We hope our findings will help practitioners working in new, understudied domains when designing deep learning systems.

\section{Diagnostic Framework: Frequency Pruning \& Spectral Behavior}
\label{sect:methodology}

\begin{wrapfigure}{r}{0.52\linewidth}
  \centering
  \vspace{-0.8\baselineskip}
  \includegraphics[width=\linewidth]{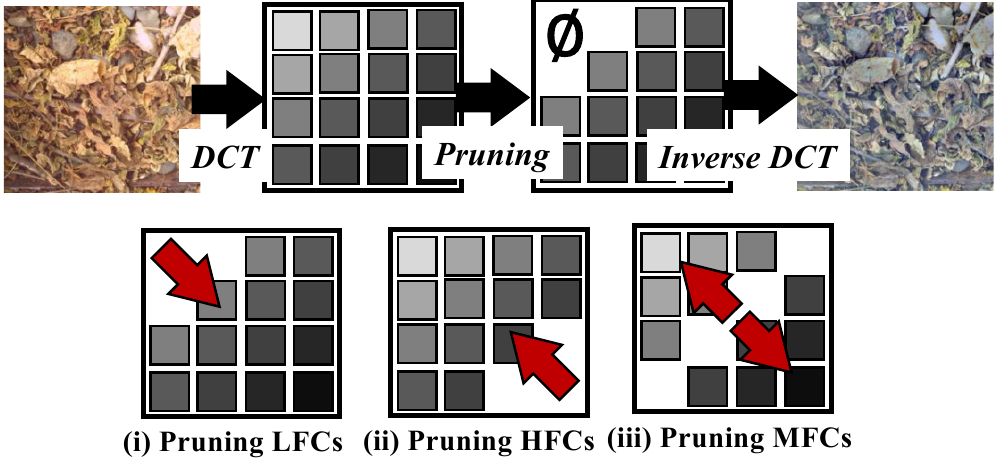}
  \caption{Frequency analysis methodology.}
  \label{fig:methodology}
  \vspace{-0.8\baselineskip}
\end{wrapfigure}

Frequency transformations provide a structured view of data \cite{jo2017}. We adopt a pruning-based analysis pipeline (Figure~\ref{fig:methodology}, top), where RGB images are transformed to the frequency domain, selectively pruned by zeroing frequency components, and inverse-transformed back to RGB for training/inference. We use the commonly used discrete cosine transform (DCT) \cite{dct,wallace_tce92,sirin2024,xu_cvpr2020}, applying it to the full image as a single block. Each color channel is transformed independently and pruned identically.

When pruning, we remove frequency components diagonally from the top-left to the bottom-right of the image (or vice versa), since oscillation rates and spatial complexity increase along this direction. Each such diagonal is referred to as a frequency component; the terms diagonal, frequency component, and component are used interchangeably. Pruning is used for \textbf{sensitivity analysis} and for \textbf{accuracy contributions}.

\textbf{Sensitivity Analysis.} We evaluate three  pruning strategies (Figure ~\ref{fig:methodology}, bottom): (i) pruning low-frequency components (LFCs), (ii) high-frequency components (HFCs), and (iii) mid-frequency components (MFCs). Each strategy removes an increasing number of diagonals in a prescribed order. Both training and test sets are pruned, and models are trained from scratch for every level of pruning. 
%, revealing which frequency types are most informative across different strategies.

\textbf{Diagonal-wise Analysis.} Existing studies either use broad units of frequency coefficients \cite{abello2021,jo2017}, or perform an analysis one coefficient at a time \cite{shortcut1}. Using broad units limits the findings, whereas one coefficient at a time analyses are too costly. Diagonal-wise grouping provides a middle ground.

\textbf{Spectral Behavior via Accuracy Contributions.} We fix the training set to either unpruned or pruned for a specific number of components, and prune test images one component at a time, attributing the resulting accuracy drop to the removed component as its accuracy contribution. The resulting distribution across components provide the spectral behavior.
%, which is a model- and task-agnostic diagnostic tool for shortcut reliance: 
%skewed distributions indicate dominance by few features, while balanced ones reflect a more robust usage. 
Distribution shifts induced by training-time interventions reflect changes in learning dynamics rather than post-hoc sensitivity.

\textbf{Low-Frequency Shortcuts.} 
We define a low-frequency shortcut as a skewed spectral behavior towards LFCs and improved ID accuracy via LFC-pruning. Our definition follows \cite{shortcut1,shortcut2} in terms of the skewed spectral behavior, but extends it by further examining the ID behavior. We do so as some domains may contain useful information within a narrow band of LFCs. \cite{geirhos2020} define shortcuts as the generalization gap between ID and OOD performance, i.e., poor OOD performance compared to the ID performance; we define low-frequency shortcuts as a broader term, including poor ID and OOD performance.

\section{Application Domains}
\label{sect:methodology_domains}

%We evaluate low-frequency shortcuts across a wide-range of texture-driven domains, spanning histopathology, material recognition, terrain classification, satellite land-cover recognition, and astronomical morphology classification.

\begin{wrapfigure}{r}{0.4\linewidth}
  \centering
  \vspace{-0.8\baselineskip}
  \includegraphics[width=\linewidth]{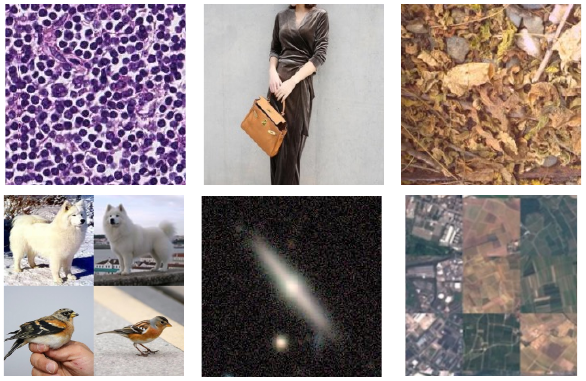}
  \caption{Sample images for the texture-driven domains we study.}
  \label{fig:example_domains}
  \vspace{-0.8\baselineskip}
\end{wrapfigure}

\noindent \textbf{Histopathology.} Histopathology analyzes microscopic tissue images for diagnosis and prognosis of diseases \cite{HMB302Inflammation,spider2025,srinidhi2021,xu2024}. Classification relies on morphological structures capturing cellular organization (1st image in Figure~\ref{fig:example_domains}).

\noindent \textbf{Textile Classification.} Driven by fashion e-commerce \cite{ecommerce}, textile classification supports applications ranging from virtual try-ons \cite{han2018} to recycling \cite{laitala2012,muthu2018}. The task requires recognizing repetitive, fine-grained patterns distributed across the image (2nd image in Figure~\ref{fig:example_domains}).

\noindent \textbf{Ground Terrain Recognition.} Ground terrain recognition supports applications such as autonomous driving \cite{shi2022,yang2024} and robot navigation \cite{guan2022,zurn2021}. Terrains correspond to surface types (e.g., leaves, grass) and are classified using surface spatial cues and texture (3rd image in Figure~\ref{fig:example_domains}).

\textbf{AI-Generated Images.} With the rapid growth of generative AI, determining whether an image is real or fake has become an important application \cite{zhu2023,park2025,ojha2023}. The task requires identifying subtle, low-level differences between real and AI-generated images of the same objects (4th image in Figure~\ref{fig:example_domains}).

\noindent \textbf{Galaxy Morphologies.} Morphological structures of galaxies are used for various purposes such as star formation. While morphologies themselves are shape-driven, the pixels creating the morphology, i.e., the stars are texture-driven \cite{galaxy_mnist} (5th image in Figure~\ref{fig:example_domains}).

\noindent \textbf{Land Use/Cover Classification.} Land use/cover classification using satellite images have numerous applications such as agriculture and urban development \cite{deepsat,landuse2010}. It includes both texture-driven, e.g., herbaceous vegetation, and shape-driven categories, e.g., high-way \cite{helber2018introducing, helber2019eurosat} (6th image in Figure~\ref{fig:example_domains}).

\section{Analysis Setup}
\label{sect:setup}

\textbf{Datasets.} We evaluate: SPIDER Colorectal histopathology  (77K samples, 14 classes) \cite{histai_spider_col,spider2025}; TextileNet (fabric subset; 350K samples, 27 classes) \cite{textilenet_github,textilenet2023}; GTOS ground terrain recognition (30K samples, 40 classes) \cite{gtos_dataset2017}; TinyGenImages (35K samples, 2 classes) \cite{zhu2023,gen_images_github,tiny_gen_images_kaggle}\footnote{Original GenImages dataset contains one million generated images across eight different generative models \cite{gen_images_github}. TinyGenImages uniformly samples 5K images across seven of the eight generative models \cite{tiny_gen_images_kaggle}. We further tested recognizing images generated by two different generative models, e.g., Midjourney to Glide, using all available images. We observed similar results. We defer these results to Appendix \ref{sect:app:midj_glide}.}; Galaxy MNIST (10K samples, 4 classes) \cite{galaxy_mnist_github}; EuroSAT (27K sampels, 10 classes) \cite{eurosat_github}; CIFAR-10 (60K samples, 10 classes) \cite{cifar10}; NICO++ domain-generalization (89K samples, 60 classes) \cite{zhang23}.
All experiments use 70\%-10\%-20\% train/val/test splits. Validation accuracies are reported. Split details are in Appendix~\ref{sect:app:setup}; test accuracies follow validation results (Appendix~\ref{sect:app:testacc}).

\textbf{Models.} We use ResNet-50 (25.6M) \cite{he2016}, MobileNet-V3 (2.5M) \cite{mobilenetv3}, ViT-Small (22M), and ViT-Tiny (5.7M) \cite{vit} as the classifiers. %\cite{vit,cifar10_resnet,deit}. 
%applying the standard CIFAR-10 architectural adaptation to preserve feature map resolution 
We train models from scratch with PyTorch’s default initialization \cite{pytorch_resnet50}. We use DinoV2 (VFM, 86M) \cite{dinov2}, DinoV3 (VFM, 840M) \cite{simeoni2026}, and CLIP (VLM, 151M) \cite{clip} as the general-purpose foundation models; and, Hibou-L (VFM, 300M) \cite{spider2025} and CONCH V1.5 (VLM, 307M) \cite{ding2025} as the domain-specific foundation models for histopathology. We train a linear layer on top of their frozen backbones. We defer further details of the foundation models to Appendix~\ref{sect:app:setup}.

% embed_dim,
% We use their standard linear-probing hyper-parameters and training regimes \cite{clip_github,dinov2_github}.

\textbf{Training.} We follow the standard CIFAR-10 training protocol for CNNs \cite{abello2021,jo2017,wang2020,shortcut1,shortcut2}; ViT \& DeiT for ViTs \cite{vit,deit}\footnote{See Table 3 of \cite{vit}, and Table 9 of \cite{deit}.}; and, linear-probing protocols for DinoV2, DinoV3, CLIP, Hibou-L, and CONCH \cite{clip_github,dinov2_github,dinov3_github,hibou_hf,conch_v15_hf}. 
Images are resized to $224{\times}224$, except EuroSAT ($64{\times}64$) and CIFAR-10 ($32{\times}32$).
Experiments are repeated with three seeds and reported as mean $\pm$ standard deviation. We report the results with seed of 42 for spectral behaviors and defer other seeds to Appendix \ref{sect:app:seeds}.
Complete training details are provided in Appendix~\ref{sect:app:setup}.

\section{Evidence of Low-Frequency Shortcuts in Texture-Driven Domains}
\label{sect:lfc_shortcuts}

\begin{figure*}[t]
  \centering
   \includegraphics[width=1\linewidth]{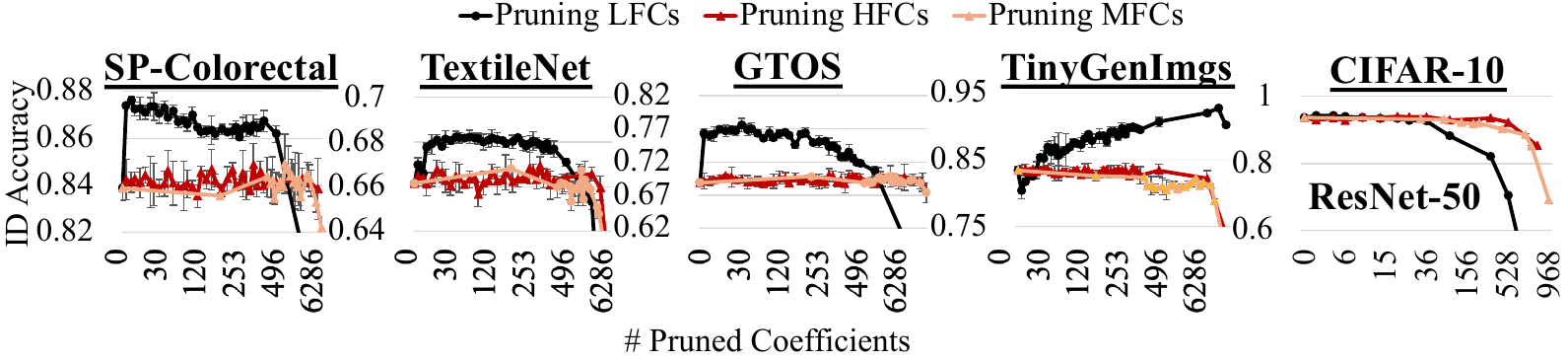}
   \caption{ID accuracy results for pruning LFCs (dark line), MFCs (light orange line), and HFCs (red line). Pruning LFCs eliminates the shortcut and improves accuracy for texture-driven tasks.}
   \label{fig:id_accuracy_resnet50}
\end{figure*}

Figure~\ref{fig:id_accuracy_resnet50} shows ID accuracy for LFC (dark), MFC (light orange), and HFC (red) pruning for the five classification tasks trained with ResNet-50. The x-axis reports the number of pruned coefficients (unscaled to capture fine-grained per-diagonal effects), and the y-axis shows ID accuracy. As can be seen, pruning HFCs and MFCs do not significantly impact the ID accuracy, as they constitute a small portion of the overall energy in images \cite{sirin2024,xu_cvpr2020,xu_phdthesis2021}.On the other hand, \textbf{pruning LFCs improves the accuracy for all the texture-driven domains}\footnote{CIFAR-10's accuracy also slightly improves (by 0.6\%), but only minimally compared to texture-driven tasks and only a little above the standard deviation (0.3\%).}.

\begin{figure*}[t]
  \centering
   \includegraphics[width=1\linewidth]{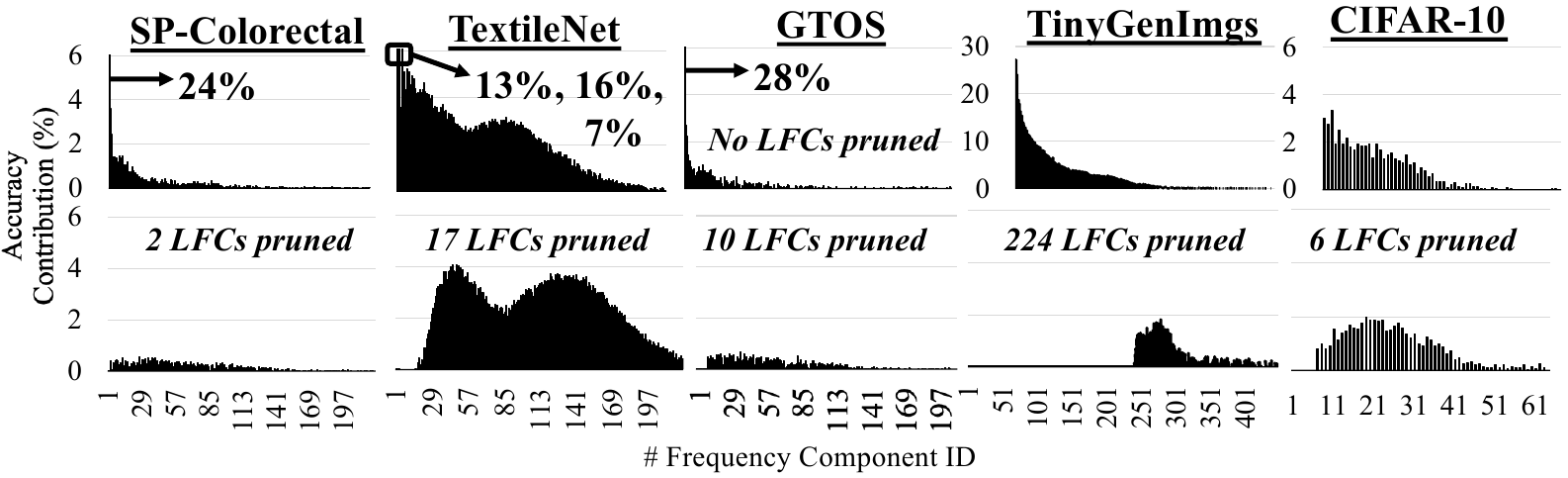}
   \caption{Accuracy contributions with unpruned (top) and pruned (bottom) training and test images.}
   \label{fig:spec_behv_resnet50}
\end{figure*}

Figure~\ref{fig:spec_behv_resnet50} presents the spectral behavior of ResNet-50 trained on unpruned images (top). Frequency component IDs increase diagonally from the top-left to the bottom-right, with accuracy contributions shown on the y-axis. As shown, \textbf{accuracy contributions are highly skewed toward LFCs} for all texture-driven domains. In contrast, CIFAR-10 exhibits a more balanced distribution. Figure \ref{fig:spec_behv_resnet50} presents the spectral behavior for pruned training images (bottom). For each task, we prune the number of LFCs that maximizes ID accuracy in Figure~\ref{fig:id_accuracy_resnet50}. As can be seen, \textbf{contributions shift toward higher frequencies} with a more balanced and unbiased distribution across all domain-specific tasks.

These results indicate that \textbf{texture-driven domains suffer from low-frequency shortcuts}. Neural networks rely exponentially more on LFCs than they do on HFCs, despite that higher-frequency features have a higher predictive power. Pruning LFCs mitigates the shortcut by shifting the spectral behavior towards higher frequencies, which in turn provides up to 10\% higher ID performance.

LFCs are background features, such as shading or camera effects. These features constitute a small set of simple features that neural networks fall into shortcut. HFCs, on the other hand, are large number of fine-grained features, and hence harder to learn from.

%We observe that texture-driven domains either learn from a few LFCs (top row of Figure \ref{fig:spec_behv_resnet50}), or a large set of HFCs (bottom row of Figure \ref{fig:spec_behv_resnet50}). Learning from a large set of features is harder than learning from a small set of features, as it requires tuning larger number of parameters. As a result, the few class-consistent low-frequency features form a shortcut. This, once again, shows that texture-driven learning is highly prone to low-frequency shortcuts, as they primarily rely on higher-frequency ranges and do not have any coherent shape-semantics that they can incorporate low and high frequencies together. 

\textbf{Prior Theoretical Work.} Simple features has been shown to have faster and stronger growing gradients than complex ones \cite{pezeshki2021,shah2020,chiang2023}. 
%based on a simple one-layer neural network and a toy dataset . Our results on real-life datasets and neural networks corroborate with the theoretical studies and reveal a novel shortcut for under-studied domains. 
We provide a detailed analysis of existing theoretical work in Appendix \ref{sect:app:theory}.

% Why are so sensitive? part 1
% 1) No shape semantics that is compatible with.
%Texture-driven domains do not have a coherent shape-semantics. As a result, any shape/low-frequency-driven feature that is consistent across the classes can easily become a shortcut and cause generalization problems. CIFAR-10, on the other hand, is shape-driven. As a result, it can incorporate both low- and high-frequency features within a coherent shape-semantics as also observed by \cite{wang2020}\footnote{See their Section 4.}. 

% TODO: 
% Footnote for other theoretical studies and refer to appendix for complete theoretical analysis reference.
% Footnote for details of the shah et al. lemma 4/5, etc.

\begin{wrapfigure}{r}{0.4\linewidth}
  \centering
  \vspace{-0.8\baselineskip}
  \includegraphics[width=\linewidth]{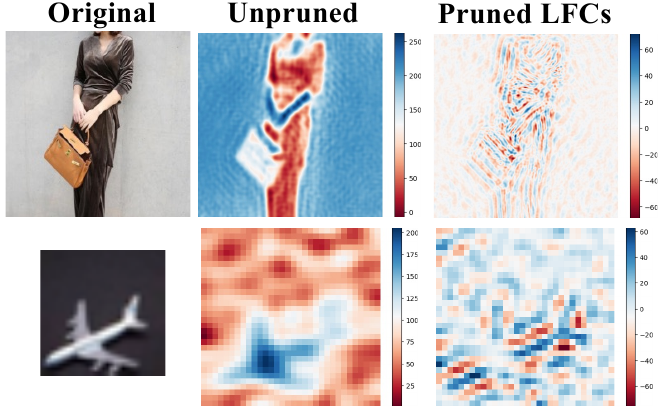}
  \caption{Sample images from TextileNet (top) and CIFAR-10 (bottom).}
  \label{fig:c10_txt_visuals}
  \vspace{-0.8\baselineskip}
\end{wrapfigure}

\textbf{Visuals.} Figure \ref{fig:c10_txt_visuals} presents a reconstructed image from TextileNet (top row) using top-contributing 50 frequency components with unpruned and pruned (17 LFCs) trainings using ResNet-50. Unpruned training learns from background and shape structures, whereas pruned training learns from fine-grained information. Further visuals are in Appendix \ref{sect:app:visuals}.

\textbf{CIFAR-10.} CIFAR-10 obtains most of its accuracy from the lower end of the frequency spectrum (Figure \ref{fig:spec_behv_resnet50}). Figure \ref{fig:c10_txt_visuals} (bottom row) presents a reconstructed sample image using top-contributing 10 frequency components with unpruned and pruned (6 LFCs) trainings. Unpruned training learns from smooth features, whereas pruned training learns from irregular structures.
%These results challenges the long-standing texture-bias hypothesis by \cite{geirhos2019} and supports the recent shape-sensitivity findings by \cite{burgert2025}. Further studying of shape/texture-bias versus low-frequency shortcuts is outside the scope of our work.

% TODO: Update summary with the new text above

\textbf{Impact of Model Architecture, Size, and Hyperparameters.} We analyzed MobileNet-V3 (2.5M params), ViT-Small (22M params), and ViT-Tiny (5.7M params), and reached similar conclusions. We further analyzed different optimizers and hyperparameters, such as learning rate, momentum, weight decay, and scheduler, for ResNet-50 and ViT-Small, and reached similar conclusions. We defer the results to Appendix \ref{sect:app:id_spec_all_models} and \ref{sect:app:hypparam}.

\textbf{Summary.} Texture-driven domains suffer from low-frequency shortcuts, where neural networks make majority of their decisions based on a few low-frequency components (LFCs), despite higher-frequency components (HFCs) have a higher predictive power. Pruning LFCs mitigate the shortcuts and provide up to 10\% improved ID accuray with a more balanced spectral behavior.

% \begin{figure*}[t]
%   \centering
%    \includegraphics[width=1\linewidth]{figures/id_accuracy_vit_small.pdf}
%    \caption{ID accuracy results for pruning LFCs (dark line) and HFCs (red line) for ViT-Small. Texture-driven tasks, SP-Colorectal and GTOS suffer from low-frequency shortcuts. Pruning LFCs eliminates the shortcut and improves ID generalization performance. CIFAR-10 is shape-driven. As a result, LFCs, defining smooth local structures, are more important than HFCs. TextileNet have a decreasing accuracy, due to the low-frequency bias of visual transformers and heavy low-frequency noises in TextileNet images.}
%    \label{fig:id_accuracy_vit_small}
% \end{figure*}

%\vspace{-0.5em}

% STRONG TODO: ADD CIFAR-10!
\section{Low-Frequency Shortcuts In OOD Classification}
\label{sect:ood}

%We study OOD classification using algorithmic corruptions by ImageNet-C benchmark \cite{hendrycks2019}, and a real-world domain-generalization dataset NICO++ \cite{zhang23}.

\subsection{Algorithmic Corruptions}
\label{sect:ood_alg}

We use ImageNet-C benchmark \cite{hendrycks2019}. We observe three main categories of corruptions in ImageNet-C: (i) low-frequency, (ii) high-frequency, and (iii) mixed corruptions \footnote{See Figure 2 of \cite{yin2019} and Figure \ref{fig:ood_freq_char} in Appendix \ref{sect:app:freq_char}.}. We select one corruption per category: fog, Gaussian blur, and elastic transform. 
We report severity 1 results with ResNet-50 and defer the rest to Appendix~\ref{sect:app:ood_severity} and \ref{sect:app:ood_all_models}.
We consider two pipelines: (i) corrupt-then-prune and (ii) prune-corrupt-prune. We report results for (i) and defer (ii) to Appendix~\ref{sect:app:ood_pipeline}. 
%Pruning test images alone filters out parts of the corruption but does not improve robustness and can substantially reduce accuracy, as observed even on uncorrupted images in accuracy contribution analyses. Pruning both training and test images instead induces a shift in the model’s spectral behavior (Figure~\ref{fig:spec_behv_resnet50}). Our goal is to characterize how this spectral shift affects OOD generalization. 

\begin{figure*}[t]
  \centering
   \includegraphics[width=1\linewidth]{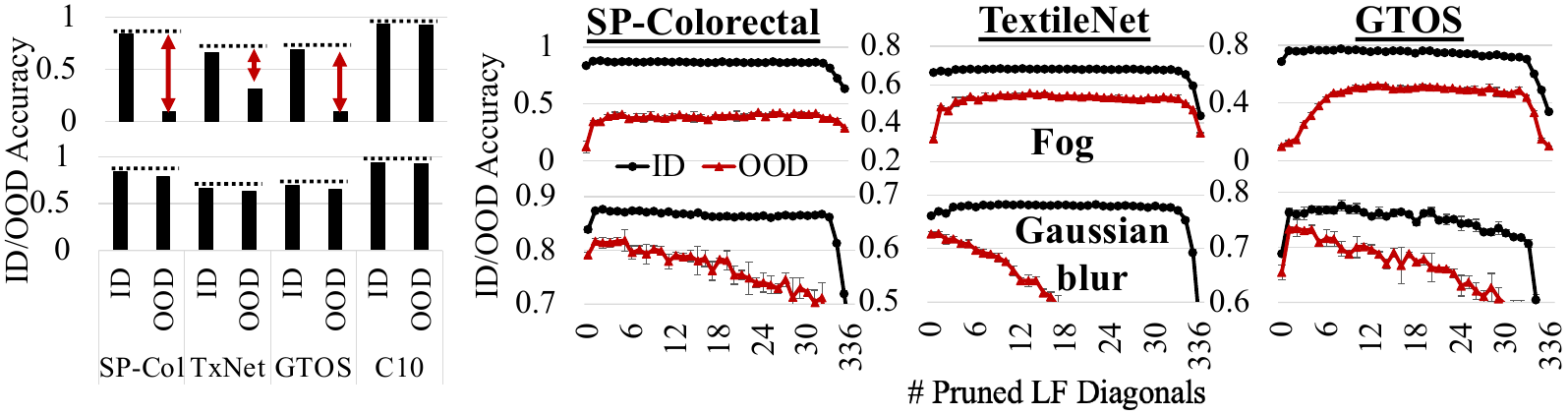}
   \caption{OOD results for fog (top) and Gaussian blur (bottom) corruptions for ResNet-50.}
   \label{fig:ood_results}
\end{figure*}

\textbf{1. Low-Frequency Corruption: Fog.}
%Fog heavily modifies low frequencies while also modifying high frequencies, but at a relatively lower scale\footnote{Figure 2 of \cite{yin2019} presents frequency characteristics in raw numbers. Hence, fog's corruptions at the HFCs are not clearly visible. We present frequency characteristics of fog, Gaussian blur, and elastic transform in log-scale in Section \ref{sect:app:freq_char} in the Appendix, where the corruptions at the higher frequencies are more clear.}. 
The left graph at top of Figure \ref{fig:ood_results} presents ID and OOD accuracies for unpruned training. OOD accuracy of texture-driven tasks heavily drops, as high as 70\%, compared to their ID accuracy. This is due to the low-frequency shortcuts, which the fog corruption heavily modifies.
The right three graphs at top of Figure \ref{fig:ood_results} present ID (dark line) and OOD (red line) accuracies. Pruning LFCs substantially improves OOD performance, by up to 40\%, thanks to eliminating the low-frequency shortcuts. 
CIFAR-10's OOD accuracy follows nearly the same curve as its ID accuracy (see Figure \ref{fig:id_accuracy_resnet50}). This is because CIFAR-10 has a more balanced spectral behavior, and hence is more robust. We defer CIFAR-10's OOD results to Appendix \ref{sect:app:cifar10_ood}.

\textbf{2. High-Frequency Corruption: Gaussian Blur.}
The left graph at bottom of Figure \ref{fig:ood_results} presents ID and OOD accuracies for unpruned training. ID and OOD accuracies are close to each other. This is because texture-driven tasks heavily rely on LFCs when trained on unpruned images. As a result, corrupting HFCs has a small impact on models' decisions. The right three graphs at bottom of Figure \ref{fig:ood_results} present ID and OOD accuracies. OOD accuracies exhibit both increasing and decreasing trends. On the one hand, generalization performance improves as HFCs better represent task semantics. On the other hand, generalization performance decreases if the corruption corrupts the HFCs that the model has started to rely on. Final accuracy depends on which factor has a greater impact. 

\textbf{3. Mixed Corruption: Elastic Transform.}
Elastic transform corrupts both low- and high-frequency components, each at lower magnitudes than fog and Gaussian blur. Its behavior is somewhere in-between fog and Gaussian blur. We defer its results to Appendix \ref{sect:app:ood_all_models}.

% \begin{figure*}[t]
%   \centering
%    \includegraphics[width=1\linewidth]{figures/nicopp.pdf}
%    \caption{Domain generalization severely suffers from low-frequency shortcuts.}
%    \label{fig:nicopp}
% \end{figure*}

\subsection{Real-World Domain-Generalization}

\begin{wrapfigure}{r}{0.47\linewidth}
  \centering
  \vspace{-0.8\baselineskip}
  \includegraphics[width=\linewidth]{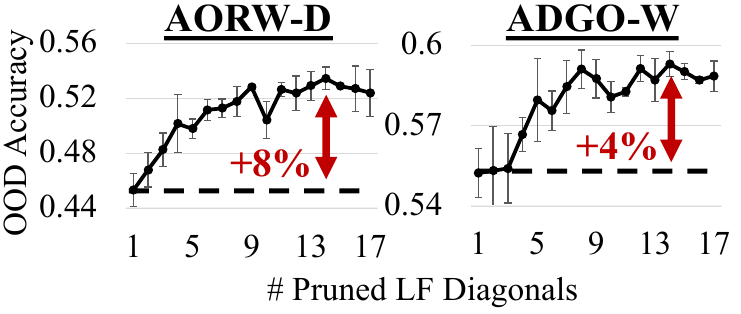}
  \caption{Domain generalization severely suffers from low-frequency shortcuts.}
  \label{fig:nicopp_small}
  \vspace{-0.8\baselineskip}
\end{wrapfigure}

NICO++ includes real-world photos of same category of images, such as rabbits or flowers, from six background domains: autumn, dim, grass, outdoor, rock, and water \cite{zhang23}. We train ResNet-50 with four domains combined and test it with one of the remaining two domains. We use initials of each domain to name each experiment. 
%AW-R, for example, stands for training the model with autumn and water, and testing it with rock.

Figure \ref{fig:nicopp_small} shows that accuracy is significantly improved, by 8\% and 4\%, as we prune LFCs. We consistently observed a similar behavior for seven other subsets of NICO++\footnote{These are AORW-G, ADGO-R, G-O, O-G, D-W, R-W, and W-D.}. 
We further observed spectral behavior of all cases are skewed towards LFCs (Appendix \ref{sect:app:nicopp_all}).
These show that low-frequency shortcuts play a crucial role in real-world domain-generalization. When there are a class-consistent low-frequency background cues, models are heavily biased towards learning from them, resulting in a highly suboptimal OOD performance. Pruning LFCs allow models to learn from more realistic higher-frequency cues with an up to 8\% improved accuracy.
%\footnote{NICO++ is a shape-driven task. On the one hand, pruning LFCs eliminates spurious background cues. On the other hand, pruning LFCs deteriorates shape structures of the objects. Forcing models to learn from smooth shape structures without relying on background cues requires further research that is outside the scope of our work.}.
We provide all results in Appendix \ref{sect:app:nicopp_all}.

\textbf{OOD Summary.} Low-frequency shortcuts play a crucial role in algorithmic as well as real-world OOD classification problems. Pruning LFCs allow models learn from realistic cues, providing up to 40\% and 8\% improved accuracy for algorithmic and real-world OOD problems.

%This shows that low-frequency shortcuts have significant impacts on OOD corruptions with losses up to 80\% compared to ID. Pruning LFCs allow models learn from a large set of higher frequencies making the models more robust, with higher ID and OOD accuracy.

% \begin{figure}[t]
%   \centering
%   \begin{tabular}{@{}m{0.38\linewidth}@{\hspace{0.04\linewidth}}m{0.58\linewidth}@{}}
%     \subcaptionbox{Spectral behavior.\label{fig:mixed_sem_spec}}%
%     {\includegraphics[width=1\linewidth]{figures/mixed_semantics_spec_behav.pdf}}
%     &
%     \subcaptionbox{ID/OOD results.\label{fig:mixed_sem_id_ood}}%
%     {\includegraphics[width=1\linewidth]{figures/new_org_mixed_semantics_id_ood_comb.pdf}}
%   \end{tabular}
%   \caption{Tasks with mixed texture and shape semantics. \textbf{(a) }Mixed-semantics tasks also suffer from low-frequency shortcuts. \textbf{(b)} Pruning LFCs results in a slight drop in ID accuracy ($\sim$1\%), but a significant increase in the OOD accuracy.}
%   \label{fig:mixed_semantics_tasks}
% \end{figure}

% TODO galaxy reconstruction is needed
% TODO galaxy seed=123/999 needed. 

% \begin{wrapfigure}{r}{0.43\linewidth}
%   \centering
%   \vspace{-0.8\baselineskip}
%   \includegraphics[width=\linewidth]{figures/mixed_semantics_spec_behav_v2.pdf}
%   \caption{Low-frequency shortcuts persist across mixed-semantics tasks.}
%   \label{fig:mixed_semantics_spec_behav}
%   \vspace{-0.8\baselineskip}
% \end{wrapfigure}

\begin{wrapfigure}{r}{0.5\linewidth}
  \centering
  \vspace{-0.8\baselineskip}
  \includegraphics[width=\linewidth]{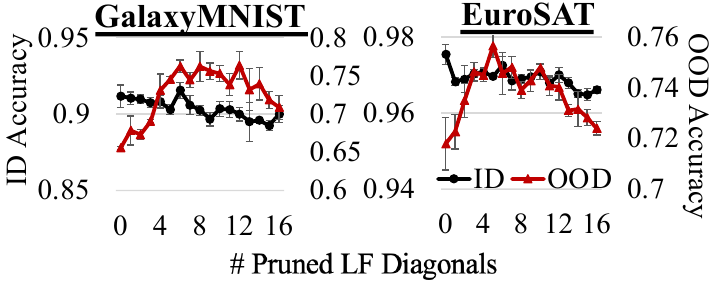}
  \caption{Skewed spectral behavior makes models vulnerable to OOD corruptions.}
  \label{fig:mixed_semantics_id_ood}
  \vspace{-0.8\baselineskip}
\end{wrapfigure}

\section{Low-frequency Shortcuts in Mixed Texture-Shape Domains}
\label{sect:mixed_semantics}

%We analyze (i) galaxy morphologies, whose classification is shape-driven (e.g., cigar vs. round-shaped galaxies), but data is texture-heavy (stars), and (ii) satellite images which has both shape (high-way) and texture-driven (forest) categories. A robust behavior in these tasks would learn from both low and high-frequency components. 

%Figure \ref{fig:mixed_semantics_spec_behav} presents the spectral behavior for GalaxyMNIST and EuroSAT. We show the lowest 16 components for brevity. Both tasks severely suffer from low-frequency shortcuts. Galaxy images could incorporate low and high-frequency features, which it fails so due to the hardship of learning from texture-driven features. EuroSAT has a class-consistent low-frequency noise (the lowest frequency component, i.e., average pixel intensity), which dominates other shape-driven, low-frequency features, once again showing \textbf{the severity of simplicity bias and low-frequency shortcuts}.

Figure \ref{fig:mixed_semantics_id_ood} presents ID (left y-axis, dark line) and OOD (right y-axis, red line) accuracy for GalaxyMNIST and EuroSAT. We report average OOD accuracy over the three corruptions used in Section \ref{sect:ood_alg}. ID accuracy drops $\sim$1\%, whereas OOD accuracy is improved by up to 3 to 7\%. Both tasks have a skewed spectral behavior towards LFCs (Appendix \ref{sect:app:mixed_semantics}). The slight drop in ID performance shows that LFCs do not constitute a shortcut for these tasks. \textbf{Nevertheless, their skewed spectral behavior make them vulnerable to OOD corruptions}, where LFC pruning significantly improves OOD performance. 

% TODO: decide what to put from below to the above. 
% A new class for example could be easily misclassified, if its underlying LFC works well but it's a slightly different galaxy. Impacting few-shot learnings, etc.  

% \begin{figure}[t]
%   \centering
%   \begin{tabular}{@{}m{0.45\linewidth}@{\hspace{0.04\linewidth}}m{0.51\linewidth}@{}}
%     \subcaptionbox{Spectral behavior.\label{fig:mixed_sem_spec}}%
%     {\includegraphics[width=1\linewidth]{figures/gtos_dino_clip_spec_behv.pdf}}
%     &
%     \subcaptionbox{aaa.\label{fig:mixed_sem_id_ood}}%
%     {\includegraphics[width=1\linewidth]{figures/gtos_dino_clip_id_ood.pdf}}
%   \end{tabular}
%   \caption{aaa.}
%   \label{fig:gtos_dino_clip}
% \end{figure}

\begin{figure*}[t]
  \centering
   \includegraphics[width=1\linewidth]{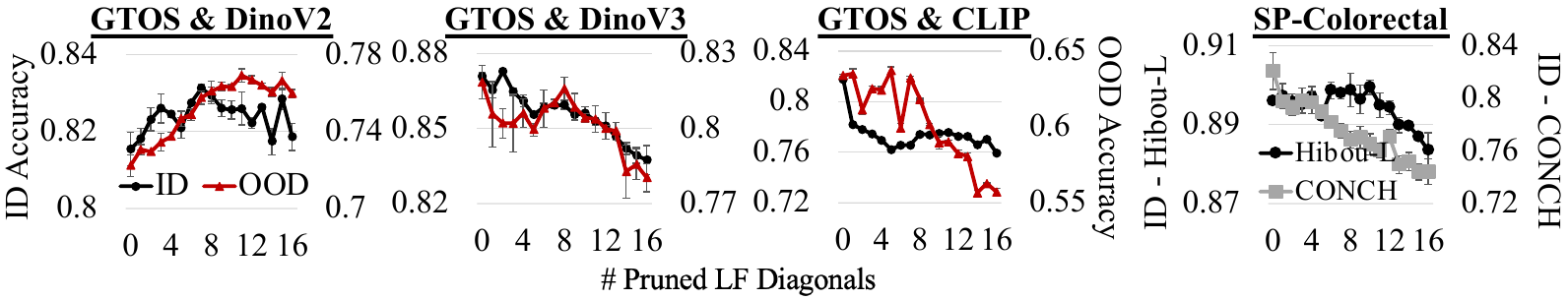}
   \caption{ID/OOD accuracy results for general-purpose (DinoV2, DinoV3, and CLIP) and domain-specific (Hibou-L and CONCH) pretrained foundation models.}
   \label{fig:all_vfms}
\end{figure*}

% \begin{wrapfigure}{r}{0.5\linewidth}
%   \centering
%   \vspace{-0.8\baselineskip}
%   \includegraphics[width=\linewidth]{figures/dinov3.pdf}
%   \caption{Large-scale Network Mitigates Shortcuts at a Cost of Large Cost.}
%   \label{fig:dinov3}
%   \vspace{-0.8\baselineskip}
% \end{wrapfigure}

\section{Impact of Large-Scale Pretraining on Low-Frequency Shortcuts}
\label{sect:vlm_vfm}

%VFMs and VLMs are pretrained on large datasets, by also using sophisticated data augmentations for VFMs. 
%We analyze a VFM, DinoV2, and a VLM, CLIP. We train a linear layer on top of a frozen backbone.

%We study low-frequency shortcuts in general-purpose and domain-specific pretraining.

\textbf{General-Purpose Pretraining.} Left three graphs Figure \ref{fig:all_vfms} present ID (left y-axis, dark line) and OOD (right y-axis, red-line) accuracy, as we prune LFCs (x-axis) for DinoV2 (86M), DinoV3 (840M), and CLIP (151M) using GTOS dataset. DinoV2's ID and OOD accuracies are significantly improved by up to 1.5\% and 4\%, respectively. 
Furthermore, DinoV2 has a skew in its spectral behavior towards first few LFCs at 6\% and 5\% (Appendix \ref{sect:app:vlm_vfm_spec_behv}). This shows that \textbf{DinoV2 suffers from low-frequency shortcuts}, even though it significantly reduces the impact of shortcuts compared to from-scratch trained ResNet-50 (third graphs in Figure \ref{fig:id_accuracy_resnet50} and \ref{fig:spec_behv_resnet50}). 
DinoV3's ID accuracy remains stable for the first few LFCs and then follows a decreasing trend. Its spectral behavior contains a slight skew towards first few LFCs at 3\% and 2.6\%. This shows that \textbf{DinoV3 mitigates the shortcuts, which, however, comes at the cost of high computational cost}, as it requires 10x more parameters than DinoV2 we use and might not be feasible for some real-world use-cases such as outdoor robotics. CLIP's ID accuracy sharply drops. Its spectral behavior contains heavy skew towards LFCs at 14\%, 8\%, 15\%, and 11\% at first four LFCs. This shows that \textbf{CLIP is heavily biased towards LFCs}. Pruning LFCs results in an information loss that CLIP cannot recover. Unlike DinoV2 and DinoV3, CLIP is not trained with complex data augmentations. Furthermore, training with text-image pairs rather than purely with images is likely to amplify the low-frequency bias, as the objective rewards broad textual descriptions of objects.

% \begin{wrapfigure}{r}{0.35\linewidth}
%   \centering
%   \vspace{-0.8\baselineskip}
%   \includegraphics[width=\linewidth]{figures/domain_spec_models.pdf}
%   \caption{Domain-Specific pretraining can mitigate the shortcuts.}
%   \label{fig:domain_spec_models}
%   \vspace{-0.8\baselineskip}
% \end{wrapfigure}

\textbf{Domain-Specific Pretraining.} Right-most graph in Figure \ref{fig:all_vfms} presents ID accuracy of Hibou-L (300M, VFM, left y-axis, dark line) and CONCH (307M, VLM, right y-axis, gray line) using SP-Colorectal dataset. Hibou-L's accuracy remains stable and then follows a decreasing trend. It spectral behavior contains a skew towards first few LFCs at 6 to 9\%. This shows that \textbf{Hibou-L mitigates the shortcuts}. While it highly relies on LFCs, it is able to recover from the lost information and maintain a similar level of accuracy as we prune LFCs. In contrast, CONCH's accuracy sharply drops, which shows that CONCH is heavily biased towards LFCs, and it cannot recover the lost information, similar to CLIP.

CONCH's accuracy (0.82) is less than from-scratch trained ResNet-50 (0.84). Furthermore, shortcut-pruned ResNet-50 achieves 0.88 accuracy, which is only 2\% less than Hibou-L (0.9), despite being >10x smaller (25M). This shows that, while domain-specific pretraining can mitigate the shortcuts, (i) \textbf{they may achieve a lower accuracy than from-scratch trained small models}, and (ii) \textbf{they are very expensive with only a few percent advantage} compared to shortcut-pruned from-scratch trained models. 

\section{Impact of Resolution, Block Size, and Color Space on Low-Frequency Shortcuts}
\label{sect:resolution_et_al}

\textbf{Resolution.} We analyze 112x112 and 56x56 images (down from 224x224) for GTOS and SP-Colorectal. We observe that \textbf{the impact of shortcuts increases as we lower the resolutions}. The accuracy contribution of the lowest-frequency component increases from 28\% to 41\% and 47\% for GTOS, and from 24\% to 33\% and 35\% for SP-Colorectal. Reducing image resolution is equivalent to concentrating information in lower-frequency bands, which results in an increased degree of low-frequency shortcuts \cite{signal_processing_book_2011}\footnote{See Chapter 5.1.3.1 and Figure 5.5 of \cite{signal_processing_book_2011}. We used PyTorch’s default bilinear interpolation and frequency-domain resizing from \cite{signal_processing_book_2011} and observed similar results.}. We provide complete results in Appendix \ref{sect:app:resolution}.

\textbf{Frequency-Transformation Block Size.} Frequency transformation is often used with various block sizes \cite{wallace_tce92, sirin2024, xu2020}. We vary the block size from 8x8, to 16x16, 32x32, and to 224x224. We perform LFC-pruning and observe ID behavior for GTOS, SP-Colorectal, and TextileNet. We observe that \textbf{large block sizes mitigate shortcuts better than small block sizes}, as they capture background cues better. We provide the results in Appendix \ref{sect:app:block_size}.

\textbf{Color Space.} We study YCrCb color space. Cr/Cb channels encode color information, and Y channel encode grayscale information. We convert original RGB images into YCrCb images, perform frequency transformation, LFC-pruning, and then convert images back to YCrCb, and then RGB. We use only Y, only CrCb, and all YCrCb channels with SP-Colorectal and GTOS datasets. We observe that \textbf{pruning LFCs from only Y or only from CrCb is not enough to eliminate the shortcuts}. Pruning LFCs from all YCrCb or RGB channels is similar. Results are in Appendix \ref{sect:app:color_spaces}. 

\section{Discussion \& Limitations}
\label{sect:discussion}

%In the long run, LFCs can be useful. But, cost budget might not allow (dinov3). Or, only a marginal improvement with a huge cost, which we haven't even used any augmentation (hibou/conch-h).

%JPEG-like usability.

%Data augmentation and compression -- 2\% difference not even augmentation. hibou-L. compression is a representation issue. 

%Methological solution -- not our goal. to shed light even beyond domains we discussed.

% ID vs. OOD difference. Optimal could be different.

% pretraining data construction structurally.

\textbf{Optimally Mitigating Low-Frequency Shortcuts.} Our work identifies a new shortcut for a class of understudied domains through extensive experimental analysis. While pruning is used as part of our diagnostic methodology, it alone does not stand for an algorithm for optimally mitigating low-frequency shortcuts. Developing such an algorithm is an exciting direction for future work. 

\textbf{Frequency-Pruning as Data Representation.} Frequency-pruned images effectively constitute a novel data representation. Section \ref{sect:vlm_vfm} showed that such representations can achieve a near-VFM accuracy. This points to an exciting future work for novel data representations including various transformations and augmentations, for an even higher accuracy at only a fraction of the cost.

\textbf{Training-Time vs. Inference-Time Images.} Images should follow a consistent data representation during training and inference. For example, if images are compressed using JPEG during training, they should also be compressed with JPEG during inference. Otherwise, the model faces a different distribution during inference than it was trained on, which can cause unpredictable performance \cite{zanjani2019,ehrlich_iccv2019,sun2018,dong2015}. 

%\textbf{Regularization \& Data Augmentation.} Pruning LFCs can be viewed as a form of regularization, where we reduce the spectral content of the input rather than constraining model parameters or capacity. \cite{yin2019} have shown that data augmentation induces a similar regularization effect. Regularization using both pruning and data augmentation is a promising future research direction. 

%\textbf{Hybrid Architectures.} We used a convolution-only or transformer-only architectures, similar to existing studies \cite{abello2021,jo2017,wang2020,shortcut1,shortcut2,yin2019}. We aim to understand learning dynamics of texture-driven tasks in basic architectural primitives, which provides the necessary foundation for more complex hybrid architectures such as MaxViT \cite{maxvit} or CoAtNet \cite{coatnet}. 

\textbf{Segmentation \& Object Detection.} We use image classification, similar to existing studies \cite{abello2021,jo2017,wang2020,shortcut1,shortcut2,yin2019}, and perform it across a wide range of domains, models, and configurations. It is not clear whether texture-driven segmentation and object detection tasks would also suffer from a similar type of shortcut. 
%For architectures that jointly optimize multiple tasks including classification, e.g., Fast R-CNN \cite{girshick2015} or Mask R-CNN \cite{he2017}, we expect our results partially hold true. 
Frequency shortcuts analysis for texture-driven segmentation and object detection tasks and architectures is future work and beyond the scope of our work.

%\textbf{ID vs. OOD Performance.} They are different. Put from rebuttal.

%\textbf{Open Questions.} Pruning is a data compression method. Our analysis shows that it can also be used for improving generalization performance. This raises a number of interesting questions. First, what is the optimal compression algorithm for a given classification task? What are the impacts of other compression primitives, such as quantization and subsampling? How do the  results change when resources are scarce? What are the ID/OOD accuracy-cost trade-offs? These are exciting research questions for future avenues of research.

% New benchmarks for OOD of application domains?
%Shape vs. texture.
% We further show that CIFAR-10 is dominated by LFCs, contributing to the recent discussion on spectral \cite{jo2017,abello2021,wang2020} and texture/shape bias \cite{burgert2025}.

\section{Related Work}
\label{sect:relwork}

Prior work primarily studies standard, shape-driven benchmarks and their OOD generalization performance. We focus on texture-driven domains and their ID and OOD generalization performance.

\noindent \textbf{Spectral Bias.} Theoretical works show that neural networks learn low-frequency functions first due to gradient-descent's simplicity bias and ReLU smoothness \cite{cao2021,dziedzic2019,lin2022,pezeshki2021,rahaman2019,basri2019,shah2020,teney2024,xu2020,xu2021,chiang2023}. Empirical works show that HFCs can be more informative than LFCs for CIFAR-10.
%, but using a broad group of HFCs \cite{abello2021,jo2017,wang2020}. 
%or both frequency ranges might matter for ImageNet \cite{shortcut1,shortcut2}. 
%Under a symmetric comparison, we find LFCs to be more important than HFCs for CIFAR-10. 
We show that, within the lower half of the spectrum, both LFCs and HFCs contribute to CIFAR-10. This is consistent with prior works, which define HFCs only broadly.
%including components from the lower half of the spectrum. 
A detailed comparison is in Appendix~\ref{sect:app:relwork}.

%Low-frequency bias arguments rely on smoothness of ReLU networks and starvation of gradients for low-energy high-frequency details. 
%Intellectually speaking, LF importance is clear by removing from training set, so there is no other way to learn. 

\noindent \textbf{Texture/Shape-Bias.} Texture/shape-bias studies show that texture/shape perturbations can both be detrimental to the model’s performance \cite{burgert2025,baker2018,gavrikov2024,geirhos2019,geirhos2020,subramanian2023,tuli2021}. Our work is orthogonal to these studies. Identifying which frequency components define what level of texture, and studying texture/shape perturbations for texture-driven domains are exciting future directions.

\noindent \textbf{Shortcut Learning.} 
%Prior work shows that models often fail to generalize even to data collected under the same protocol as the original dataset \cite{recht2019}. 
To mitigate shortcuts, existing approaches \cite{wang2024} include feature-space control  \cite{cleverHans2019,niu2022,pezeshki2021,zheng2022}, data augmentation \cite{geirhos2019,minderer2020,nauta2022,xu2025,yin2019}, layer-wise penalization \cite{boland2024,wang2019}. 
Our work is orthogonal to these studies, as it takes the first step and performs an extensive experimental analysis to identify a new shortcut for a new class of domains. Proposing a novel algorithm to optimally mitigate low-frequency shortcuts for texture-driven domains is an exciting future work.

\section{Conclusion}
\label{sect:conclusion}

We show that texture-driven domains suffer from low-frequency shortcuts, where a small number of low-frequency components (LFCs) dominate the model's decisions with a skewed spectral behavior, despite that higher-frequency components (HFCs) have a higher predictive power. Pruning LFCs shifts the spectral behavior towards higher frequencies, which in turn provides a significantly improved ID and OOD accuracy, by up to 10\% and 40\%, respectively. We show that low-frequency shortcuts persist across different model sizes, architectures, and hyperparameters, under algorithmic as well as real-world OOD conditions, for various pretrained foundation models, as well as different image resolutions, frequency-transformation block sizes, and color spaces. Our study contains valuable insights that we hope will be useful to practitioners working in new, understudied domains.

\noindent \textbf{Acknowledgments.} DAM acknowledges support from the Kempner Institute, FAS Dean’s Competitive Fund for Promising Scholarship, Aramont Fellowship Fund, and the NSF AI-SDM Institute (Grant No. IIS-2229881).

\newpage

\appendix

\setlength{\tabcolsep}{4pt} % default is 6pt
\renewcommand{\arraystretch}{1.05}
\begin{table}[t]
\centering
\caption{Datasets.}
\label{tab:datasets}
\begin{tabular}{l c c c}
\toprule
Dataset & \#Classes & Train / Val / Test & Input Size \\
\midrule
CIFAR-10 & 10 & 45K / 5K / 10K & $32\times32$ \\
SP-Colorectal & 14 & 56K / 8K / 13K & $224\times224$ \\
TextileNet & 27 & 245K / 35K / 70K & $224\times224$ \\
GTOS & 40 & 24K / 2K / 4K & $224\times224$ \\
TinyGenImages & 2 & 24K / 7K / 4K & $224\times224$ \\ 
GalaxyMNIST & 4 & 8K / 1K / 1K & $224\times224$ \\
EuroSAT & 10 & 19K / 2.7K / 5.4K & $64\times64$ \\
NICO++ & 60 & 64.1K / 17.8K / 7.1K & $224\times224$ \\
\bottomrule
\end{tabular}
\end{table}

\section{Additional Experimental Details}
\label{sect:app:setup}

\noindent \textbf{Datasets}. We use one image classification dataset for each domain we described in Section \ref{sect:methodology_domains}. Table \ref{tab:datasets} presents the summary of each dataset. For histopathology, we use the recently proposed multi-organ SPIDER dataset \cite{spider2025}. We use the colorectal organ type, due to its widespread adoption \cite{histai_spider_col}. It is a large-scale dataset with $\sim$77K images and 14 classes. We use only the central patches. We use the original training-test set split with 80\%-20\% ratio. We split the training set further into 70\% training and 30\% validation. We name this dataset as \textbf{SP-Colorectal} in our figures. For the textile classification dataset, we use the recently proposed \textbf{TextileNet} dataset \cite{textilenet2023}. We use its fabric subset, which contains 27 classes and $\sim$350K samples \cite{textilenet_github}. We use the proposed 70\%-10\%-20\% training-validation-test set splits \cite{textilenet2023}. For ground terrain recognition, we use the recent ground terrain dataset for outdoor scenes, \textbf{GTOS} \cite{gtos_dataset2017}. It is a large-scale dataset with $\sim$30K images and 40 classes. We use the first of its five original 70\%-30\% training-test splits. We further split the test set into validation and test sets by 10\%-20\%. For AI-generated images, we use \textbf{GenImages} and \textbf{TinyGenImages} datasets \cite{gen_images_github, tiny_gen_images_kaggle, zhu2023}. It includes original images taken from the ImageNet dataset \cite{imagenet}, and their corresponding fake images generated by eight different generative models. TinyGenImages is a uniformly sampled version of GenImages. GenImages includes one million generated and one million original images. TinyGenImages includes 17.5K generated and 17.5K original images. We use the original training and test split (28K / 7K), where we split the training set further into training and validation sets of 85\% to 15\% (size 24K and 4K).
We use the basic \textbf{Galaxy MNIST} dataset with 10K samples, 4 classes \cite{galaxy_mnist_github} among others \cite{galaxy_zoo,galaxiesml} due to its convenience. We use the original train/test split, and further split the training set into train/val split by 90-10\%. We use their 224x224 version (GalaxyMNISTHighrez).
We use \textbf{EuroSAT} dataset for satellite images with 27K samples and 10 classes covering shape- and texture-driven categories \cite{eurosat_github}. We randomly split the dataset into 70-10-20\% train/val/test splits, as there is no original train/val/test splits.
As the standard benchmark, we use \textbf{CIFAR-10} \cite{cifar10} due to its widespread adoption in frequency characterization studies \cite{abello2021,jo2017,wang2020}. It has 50K training and 10K test images over 10 classes. We split the training set into 45K-5K training-validation sets. \textbf{NICO++} is a domain-generalization dataset, where there are 60 object categories taken in six different domains, such as rock, water, autumn, dim, water, and grass\footnote{While the paper refers to 10 domains with 80 categories each \cite{zhang23}, the published dataset contains a subset of it with six domains and 60 categories \cite{nicopp_github}.}. It has 89K samples. We use the original training and test splits of 80-20\%, which then we split the training set into further training and validation of 90-10\% \cite{nicopp_github,zhang23}.

\textbf{Foundation Models.} For DinoV2, DinoV3, and CLIP, we train a linear layer (768, 1280, and 512 units, respectively) on top of their frozen backbones. We use ViT-B/14-Reg (86M params), ViT-H+/16 (840M params) and ViT-B/32 (151M params) backbones, respectively. For CONCH V1.5 and Hibou-L, we train a linear layer (768 and 128 units, respectively) on top of their frozen backbones. We use ViT-L/16 (307 params) and ViT-L/14 (300M params) backbones, respectively. 

\noindent \textbf{Training.} 
We train with SGD (momentum 0.9, weight decay $5{\times}10^{-4}$), a cosine annealing schedule with initial learning rate 0.1, and batch size 128 for 200 epochs, selecting the checkpoint with highest validation accuracy for ResNet-50 and MobileNet-V3. Following \cite{vit,deit}, we use AdamW (weight decay=0.3) with initial learning rate 3e-3, 300 epochs of training, 3.4$\times$(number of batches) warmup steps, using linear scheduler for the warmup (start factor 1e-8, end factor 1.0), and cosine scheduler for the post-warmup training, batch size of 512.
We train a linear head on top of a frozen DinoV2 and CLIP backbones (768 \& 512 units, respectively) for 100 epochs with a batch size of 32. We use lr=$10^{-4}$, weight decay=$10^{-4}$ for DinoV2, and lr $10^{-3}$ and weight decay $10^{-4}$ for CLIP. We use random-resized-crop and horizontal-flip augmentations for DinoV2, and no data augmentation for CLIP, following the original code bases for linear probing \cite{dinov2_github, clip_github} 
We use seed values of 42, 123, and 999. We make sure we achieve, for each benchmark, an accuracy almost as high as in their original papers. This is 89\% for SP-Colorectal, 66\% for TextileNet, 77\% for GTOS, and 94\% for CIFAR-10. 
We use training-validation-test splits, as explained earlier in this section, and report test-set accuracies in Appendix \ref{sect:app:testacc}. Validation-test set splits, and their accordance are important to keep relevance to image compression studies, which need to perform validation set analysis to decide how many images should be compressed at test time \cite{sirin2024,xu_cvpr2020}. We normalize each image using the mean and standard deviation of its training set, separately for each pruned version of the training set. 

%Interestingly, maximum accuracy results that we observe for each benchmark is similar to the reported accuracies in their original papers. Our data pruning improves accuracy just like what original papers do in their original papers like with data augmentation for sp-colorectal and architectural stuff for GTOS. Great observation, defintely reveal it!

\noindent \textbf{Hardware}. We use a cluster of Nvidia A100, H100, and H200 GPUs. We submit jobs as batches and use GPUs as they are available. Each training takes 1-6 hours with a single GPU. Obtaining the spectral behavior requires an absence-test per diagonal, for each diagonal of coefficients. Each such experiment requires one pass over the validation set, which is usually a few minutes on a single GPU. OOD tests are similar to spectral behavior tests, as they also require one pass over the corrupted validation images. 

\begin{figure*}[!ht]
  \centering
   \includegraphics[width=1\linewidth]{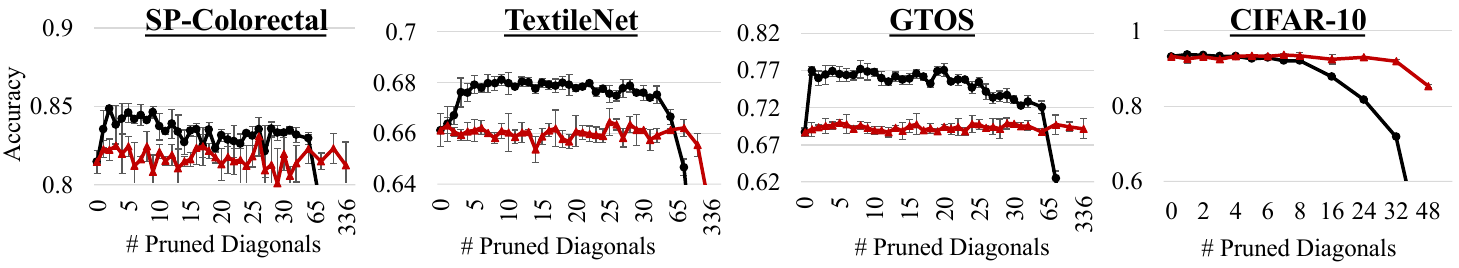}
   \caption{Test set ID accuracy results for pruning LFCs (dark line) and HFCs (red line). Test-set and validation set results are similar. Pruning LFCs improves accuracy for texture-driven tasks by improving representation and reducing frequency shortcuts. LFCs are crucial for CIFAR-10, as it is shape-driven and depends on smooth, high-level structures. Pruning HFCs minimally impacts accuracy, especially for SP-Colorectal and GTOS. TextileNet and CIFAR-10 are more dependent on the higher frequencies at the lower half of the spectrum. Hence, they suffer from observable accuracy losses near the end of HFC pruning.}
   \label{fig:id_accuracy_test_set}
\end{figure*}

\begin{figure*}[!ht]
  \centering
   \includegraphics[width=1\linewidth]{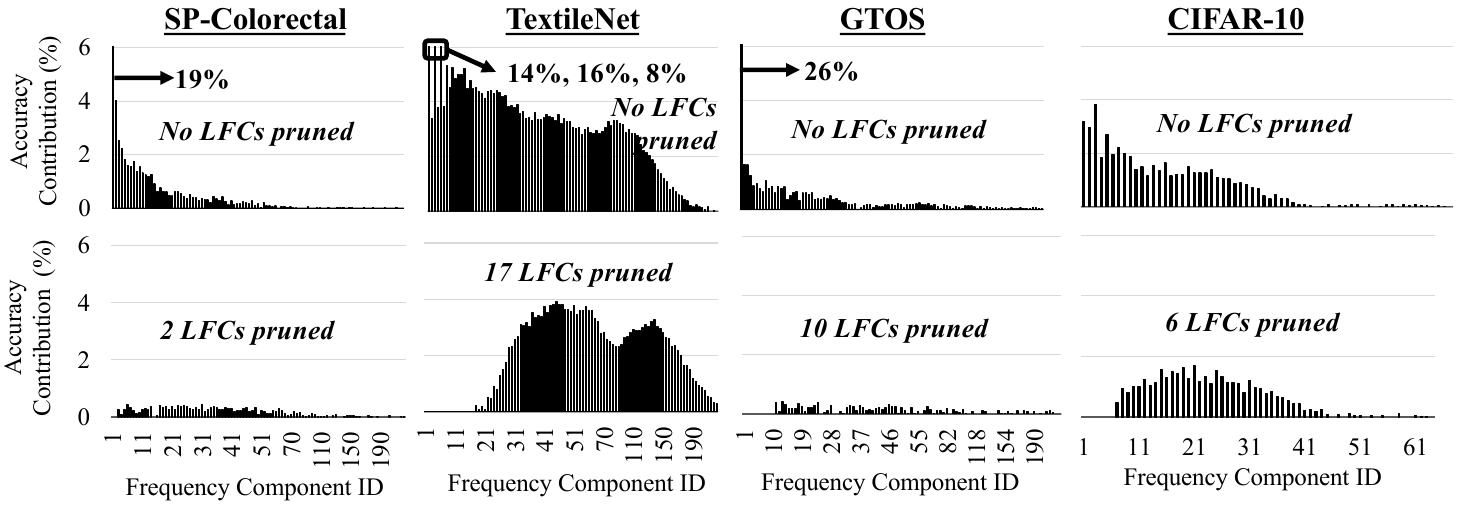}
   \caption{Accuracy contributions based on test-set images. \textbf{Top:} Accuracy contributions when models are trained with unpruned images. \textbf{Bottom:} Accuracy contributions when models are trained with LFC-pruned images. Results follow similar patterns to the validation-set results. Unpruned training  causes a skewed behavior for texture-driven tasks, indicating a low-frequency shortcut. LFC-pruned training eliminates the shortcut and improves representation capacity. CIFAR-10 benefits from a range of components from the lower end of the frequency spectrum.}
   \label{fig:shortcut_all_test_set}
\end{figure*}

\begin{figure*}[!ht]
  \centering
   \includegraphics[width=1\linewidth]{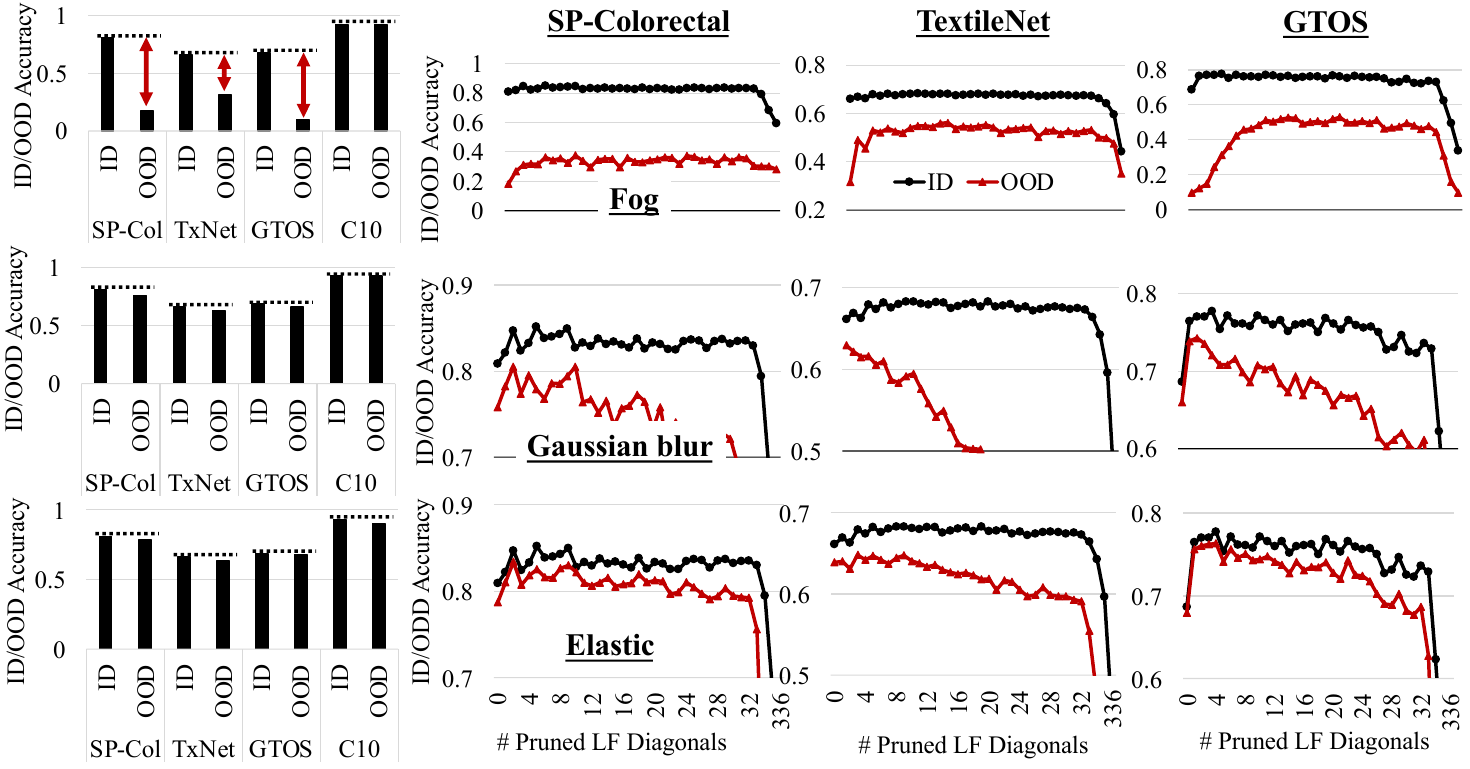}
   \caption{OOD corruption results based on test-set images. \textbf{Top:} Low-frequency corruption (fog) results. \textbf{Middle:} High-frequency corruption (Gaussian blur) results. \textbf{Bottom:} Mixed-frequency corruption (elastic transform) results. Results are similar to those of validation-set images. Fog heavily impacts texture-driven tasks due to the low-frequency shortcuts. Pruning LFCs significantly improves OOD accuracy for fog corruption. Pruning LFCs introduces a trade-off under Gaussian blur and elastic transform corruptions: OOD accuracy improves due to HFCs' improved representation, but decreases due to magnified corruptions at the HFCs. Final accuracy depends on which factor has a greater impact.}
   \label{fig:ood_accuracy_test_set}
\end{figure*}

\section{Test-Set Results}
\label{sect:app:testacc}

This section presents test set results for ID performance, accuracy contributions, and OOD performance. Figure \ref{fig:id_accuracy_test_set} presents test set results for ID performance. As can be seen, results follow similar trends to the validation set accuracies. Texture-driven tasks, SP-Colorectal, TextileNet, and GTOS suffer from low-frequency shortcuts. Pruning LFCs eliminates the shortcut and improves ID generalization performance. CIFAR-10 is shape-driven. As a result, LFCs, defining smooth local structures, are more important than HFCs. HFCs are promising candidates for image compression, as pruning up to 87.5\% of them results in no loss of accuracy.

Figure \ref{fig:shortcut_all_test_set} presents accuracy contributions based on test set images. As shown, the results are similar to those of the validation set. All texture-driven domains suffer from low-frequency shortcuts (top row), with a skewed distribution towards LFCs (top row). When models are trained on pruned images (bottom row), the spectral behavior of the texture-driven tasks shifts towards higher frequencies, with a more balanced distribution. CIFAR-10 has a significantly more balanced distribution than texture-driven tasks. CIFAR-10's distribution also shifts toward higher frequencies, although largely remaining within the lower half of the spectrum. 

Figure \ref{fig:ood_accuracy_test_set} presents test set results for OOD corruptions. The results closely follow the validation-set results. Low-frequency corruptions (fog) heavily impact texture-driven domains, due to the low-frequency shortcuts. Pruning LFCs eliminates the shortcut, resulting in significantly higher OOD accuracy against fog corruption. Nevertheless, the maximum OOD accuracy is significantly lower than the maximum ID accuracy, due to the magnified corruption at HFCs. Fog, despite being low-frequency-heavy, also corrupts HFCs. When models are trained with LFC-pruned images, they rely more on HFCs and LFCs. As a result, they become more sensitive to high-frequency corruptions, and fog's corruptions at the HFCs are magnified. As a result, final OOD accuracy is significantly lower than the ID accuracy.

Gaussian blur and elastic transform minimally affect ID accuracy, as their corruption at the LFCs is not significant. As LFCs are pruned, models latch onto higher frequencies that better represent application semantics, while becoming increasingly vulnerable to high-frequency corruptions. Hence, OOD accuracy increases, stays stable, or decreases depending on whether the improved representation or magnified corruption is more significant.

\begin{figure}[!ht]
  \centering
   \includegraphics[width=1\linewidth]{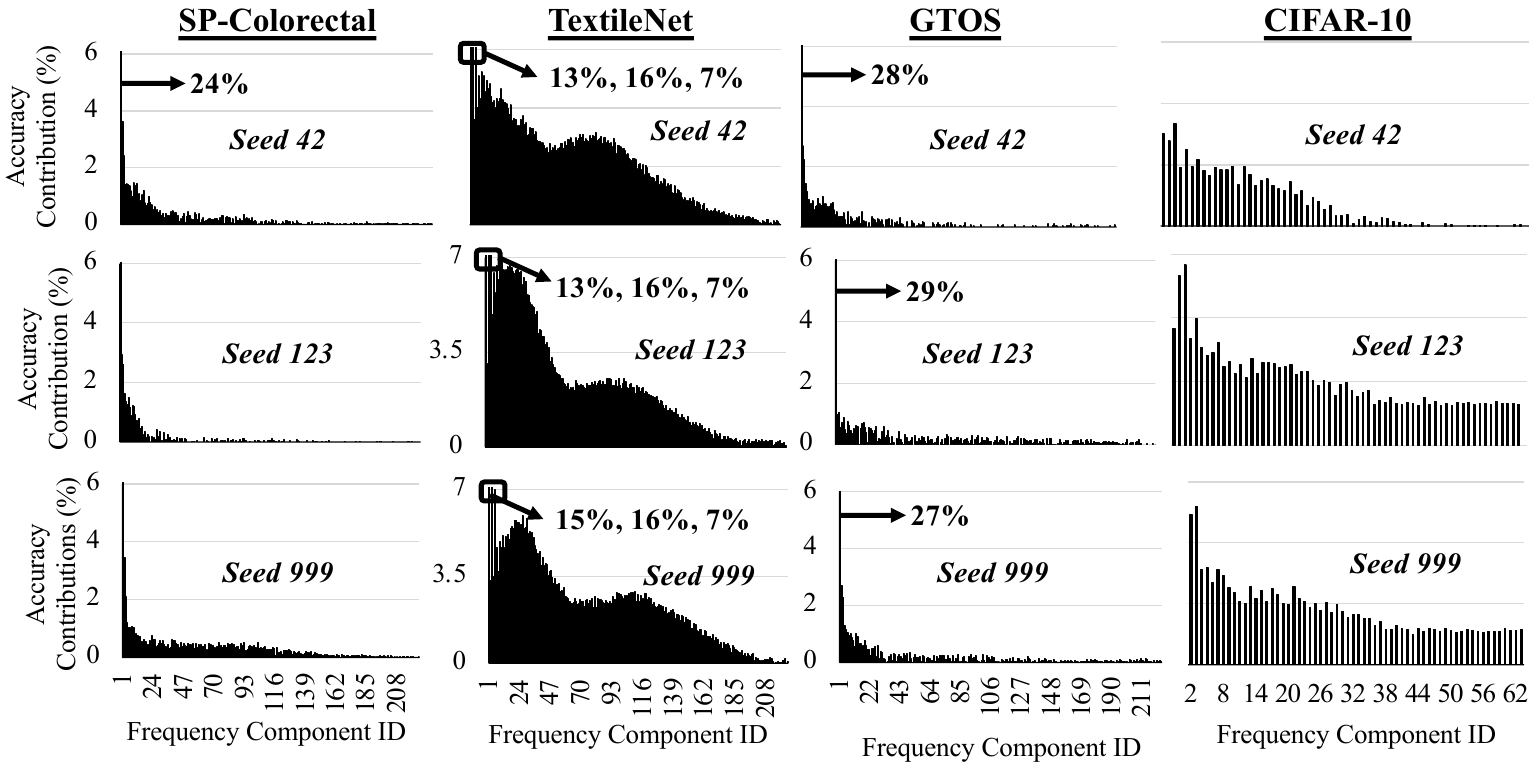}
   \caption{Spectral behavior for ResNet-50 across different seeds when using unpruned images. As can be seen, similar trends are observed across different seeds.}
   \label{fig:spec_behv_seeds_unpruned}
\end{figure}

\begin{figure}[!ht]
  \centering
   \includegraphics[width=1\linewidth]{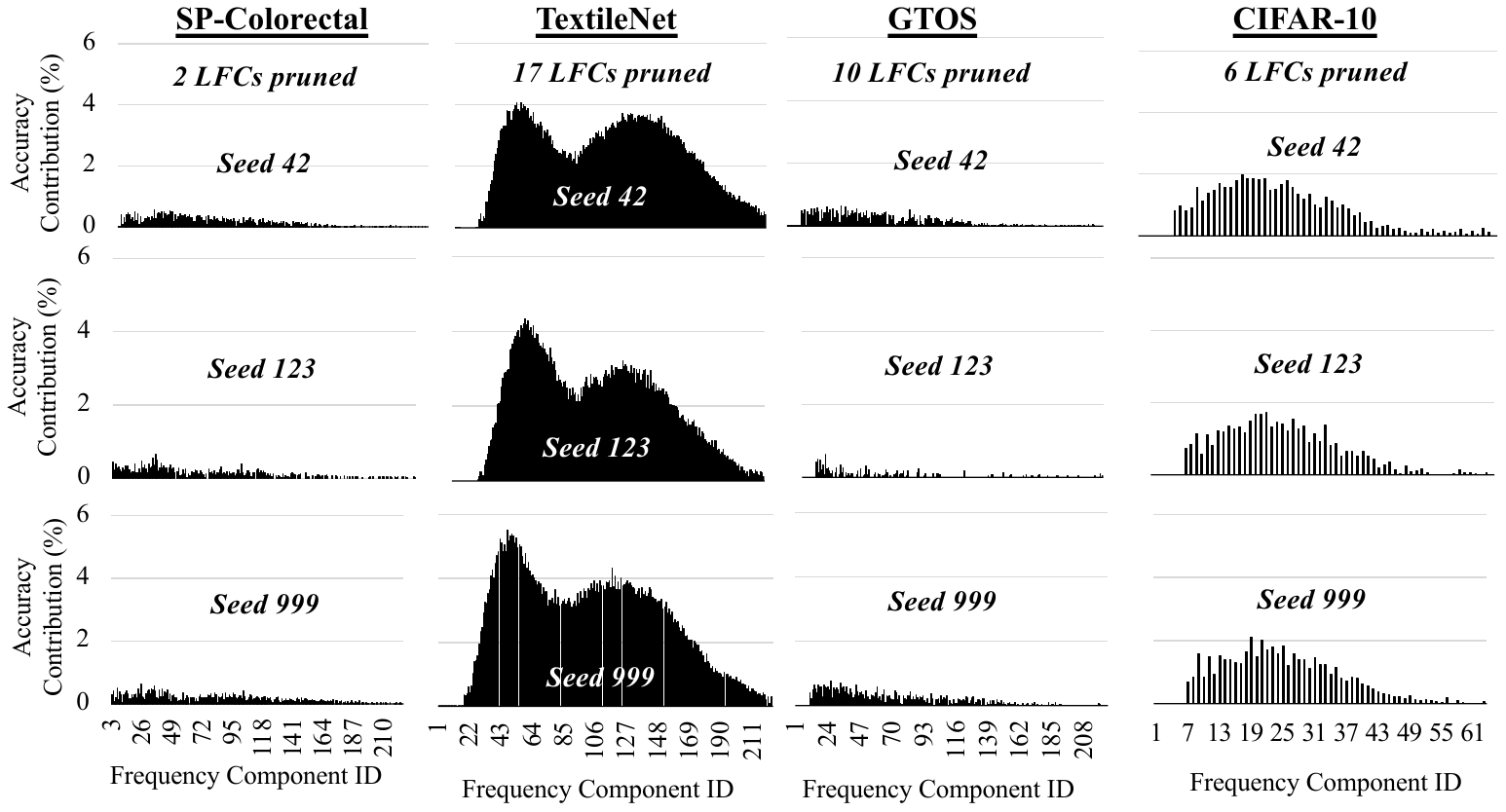}
   \caption{Spectral behavior for ResNet-50 across different seeds when using pruned images. As can be seen, similar trends are observed across different seeds.}
   \label{fig:spec_behv_seeds_pruned}
\end{figure}

\section{Spectral Behavior Across Different Seeds}
\label{sect:app:seeds}

We report spectral behavior, i.e., the accuracy contributions for ResNet-50 when trained with unpruned images for SP-Colorectal, TextileNet, GTOS, and CIFAR-10 in Figure \ref{fig:spec_behv_seeds_unpruned}. As can be seen, similar trends are observed across different seeds. Similarly, Figure \ref{fig:spec_behv_seeds_pruned} presents the spectral behavior for pruned images, where we prune as many LFCs as each model achieves its maximum ID accuracy. Once again, the figure shows that the trends holds true across different repetitions. This shows that low-frequency shortcuts are persistent across different seeds and repetitions.

\section{Theoretical Discussion}
\label{sect:app:theory}

Our results closely follow the theoretical results presented by Shah et al. 2020 \cite{shah2020}, Pezeshki et al. 2021 \cite{pezeshki2021}, and Chiang et al. 2023 \cite{chiang2023}. 

Shah et al. show that neural networks are provably biased towards learning simpler decision boundaries (see their Figure 1, green line) than complex decision boundaries (see their Figure 1, orange line). The authors prove that the gradients for the simpler decision boundary are consistently larger than those for the complex decision boundary (see Section 4.1, Theorem 1). The proof stems from the closed-form gradient expressions for the simple and complex decision boundaries given by Lemmas 4 and 5 in Appendix F.2, respectively. 

The first lines of the proof of each lemma have the $y_jx_{1j}$ and $y_jx_{2j}$ terms in them, which is the reason that the two gradients have different final expressions. $y$ stands for the label, $x_{1j}$ stands for the feature value providing a simple decision boundary for sample $j$, and $x_{2j}$ stands for the feature value providing a complex decision boundary for sample $j$. $y_jx_{1j}$ always produces the same result for all the samples (1 in the specific scenario), whereas $y_jx_{2j}$ produces different results depending on the feature and label values (1, -1, or 0 in the specific scenario). This results in constant increases in the gradients of the parameters for the simple feature ($\nabla_{w_{1i}}$, where $i$ is the neuron index), whereas gradients of the parameters for the complex feature ($\nabla_{w_{2j}}$ fluctuate (in a one-layer neural network). As a result, parameter values of the simple feature (Linear Coordinate in Theorem 1) are constantly bigger than the parameter values for the complex feature (3-Slab Coordinate in Theorem 1). The neural network then relies on the feature that can be used with a simpler decision boundary to make most of its decisions. The authors show that simple features are chosen over complex features, even when simple features have a less predictive power, indicating the severity of the simplicity bias (see their Section 5).

Pezeshki et al. introduce the concept of Gradient Starvation that supports the findings of Shah et al. The authors use a simplified Neural Tangent Kernel regime, where the output of the network can be approximated as a linear function. The authors use SVD to decompose the label-feature matrix ($Y\Phi_0$) into its principal components and define the strength of a feature by its corresponding eigenvalue. Features that are highly aligned with the labels (e.g., the horizontal feature in their Figure 1) have large eigenvalues, and vice versa.  

The authors show that if a feature has a high strength, i.e., is highly correlated with the label, i.e., simple in Shah et al.’s definition, the strength of the feature decreases the response of the network to the other feature (see Theorem 2, Eq. 16 ($\frac{dz^*_2}{ds^2_1}<0$) where $z^*_2$ is the optimal response of the network to feature 2, and $s^2_1$ is the square of the strength of feature 1), which the authors define as the Gradient Starvation (although they do not explicitly show that gradients diminish over the iterations). 

\begin{wrapfigure}{r}{0.38\linewidth}
  \centering
  \vspace{-0.8\baselineskip}
  \includegraphics[width=\linewidth]{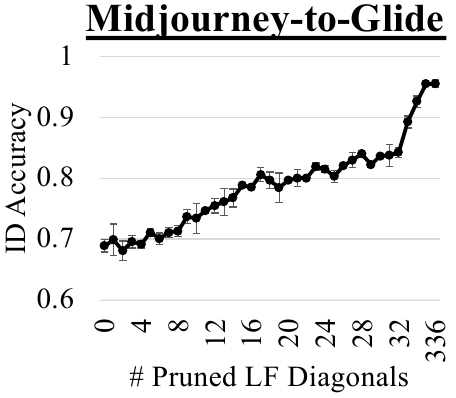}
  \caption{ID results for a subset of original GenImages dataset.}
  \label{fig:midj_glide}
  \vspace{-0.8\baselineskip}
\end{wrapfigure}

Finally, Chiang et al. observe that simple solutions are favored by non-gradient-based optimizers as well and attribute this to the volume hypothesis, where simple solutions occupy orders of magnitude more volume in the loss landscape than complex solutions. The authors show that the simple solution has a volume that is 6 orders of magnitude larger than the complex solution for the same example Shah et al. uses (see Chiang et al.’s Section 6 \& Figure 3).

These theoretical results explain what we empirically observe at scale: neural networks learn from a few LFCs rather than a large number of HFCs in texture-driven domains, despite the fact that HFCs have greater predictive power. We observe that neural networks either (i) learn from a few LFCs or (ii) a large number of HFCs. It is never that the network learns from a small number of HFCs. This shows that LFCs provide a simple set of features that are highly correlated with the labels, similar to the simple features that Shah et al. studied in their examples. Once the network starts learning from a few LFCs, the gradients of the LFCs starve the gradients of the HFCs, and eventually, the LFCs dominate the learning process. When we prune the LFCs from the training set, the network can learn from a complex set of HFCs, as there is no other competitor now. 

\section{Additional Results on AI-Generated Images}
\label{sect:app:midj_glide}

In addition to the TinyGenImages results in Section \ref{sect:lfc_shortcuts}, we analyze a specific subset of original GenImages \cite{gen_images_github,zhu2023} to make sure TinyGenImages' sampled images does not introduce any misleading outcome for our analyses. We observe that the original GenImages dataset contains $\sim$160K generated images for each generative model, across eight different generative models, such as BigGAN, Midjourney, and Glide. Table 2 of \cite{zhu2023} shows that main challenge of the GenImages dataset is to accurately predict images generated by two different generative models. Hence, we randomly choose two different generative models and analyze its ID accuracy behavior as we prune LFCs. We choose Midjourney and Glide. We train the model (ResNet-50) using Midjourney-generated images and test it with Glide-generated images. 

Figure \ref{fig:midj_glide} presents the results. As can be seen, accuracy constantly improves by up to 25\%, even more than that of TinyGenImages shown in Figure \ref{fig:id_accuracy_resnet50} in Section \ref{sect:lfc_shortcuts}. This shows that our results with TinyGenImages is representative for full GenImages.

\begin{figure}[!ht]
  \centering
  \begin{subfigure}{0.48\linewidth}
    \centering
    \includegraphics[width=\linewidth]{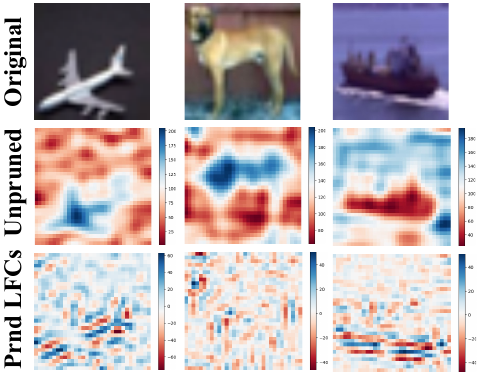}
    \caption{CIFAR-10 visualizations. Models trained with unpruned images latch onto the smooth structures that define object shapes well. Models trained with pruned images (6 LFCs) latch onto the more irregular structures. As we prune a larger and larger number of LFCs, shapes become increasingly distorted, and accuracy drops sharply.}
    \label{fig:app_visuals_shape}
  \end{subfigure}\hfill
  \begin{subfigure}{0.48\linewidth}
    \centering
    \includegraphics[width=\linewidth]{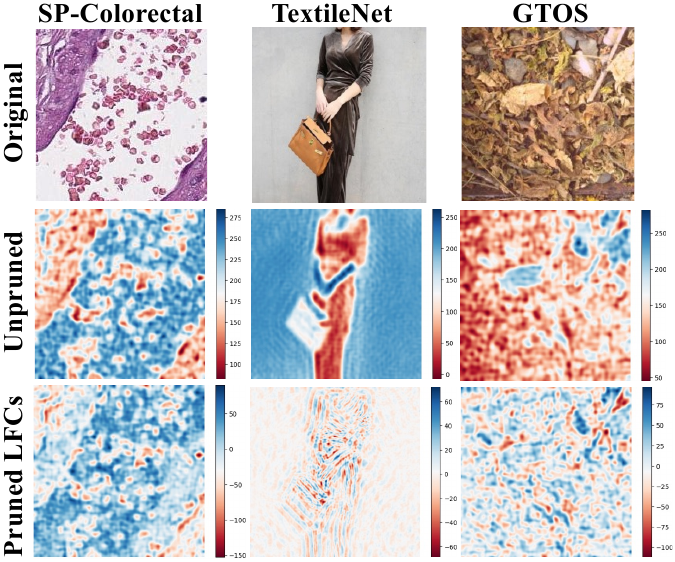}
    \caption{Texture-driven tasks' visualizations. Models trained with unpruned images latch onto large blobs and shapes that do not have any meaningful, low-level structure that matches with the application semantics. Models trained with pruned images latch onto fine-grained details that defines the application semantics more faithfully.}
    \label{fig:app_visuals_texture}
  \end{subfigure}
  \caption{Sample images from CIFAR-10 (left) and texture-driven tasks (right). \textbf{Top:} Original images. \textbf{Middle:} Reconstructed images based on unpruned training. \textbf{Bottom:} Reconstructed images based on pruned training.}
  \label{fig:app_visuals_combined}
\end{figure}

\section{Additional Visualizations}
\label{sect:app:visuals}

This section presents additional visualizations for CIFAR-10 and the texture-driven domains we studied. 
Figure \ref{fig:app_visuals_shape} presents sample images for CIFAR-10. Images are reconstructed by using the top 10 frequency components in terms of their accuracy contributions in Figure \ref{fig:spec_behv_resnet50}. The top row presents the original images from the airline, dog, and ship classes, the middle row presents the reconstructed images from unpruned training, and the bottom row presents the reconstructed images from 6-LFC-pruned training. Our goal is to visualize the data characteristics that neural networks learn from when they are trained with unpruned versus pruned images. We use colormaps because we want to focus on the strength of the signal components and their characteristics.

As can be seen, models trained on unpruned images (middle row) latch onto structures with smooth patterns that define the overall shape of the object, which are largely LFCs, as shown in the accuracy contributions in Figure \ref{fig:spec_behv_resnet50}. When models are trained on LFC-pruned images, they latch onto higher frequencies that still include a significant amount of information, but are much more irregular and significantly distort the overall shape of the object. This explains both how accuracy is preserved when LFCs are pruned and how it starts to drop significantly after pruning 6 LFCs. Up to 6 LFCs; while patterns start breaking, they still contain sufficient information to perform classification. As we prune an increasing number of LFCs, the patterns become increasingly distorted, resulting in a significant loss of accuracy. 

Figure \ref{fig:app_visuals_texture} presents sample images reconstructed from frequency components with the top 50 frequency components in terms of their accuracy contributions shown in Figure \ref{fig:spec_behv_resnet50}, for the texture-driven tasks. The top row presents original images; the middle row, images reconstructed from models trained on unpruned images; and the bottom row, images reconstructed from models trained on pruned images. When pruning, we prune as many LFCs as needed to achieve the highest ID accuracy. 

As can be seen, models trained on unpruned images (middle row) latch onto low-frequency information that largely defines broad shapes and blobs, with minimal fine-grained structure. On the other hand, models trained on LFC-pruned images (bottom row) latch onto fine-grained, high-frequency structures that are better aligned with the characteristics of the classification tasks. This allows learning from a range of higher-frequency components with a more balanced, unbiased distribution, resulting in better generalization performance.

\begin{figure}[!ht]
  \centering
   \includegraphics[width=1\linewidth]{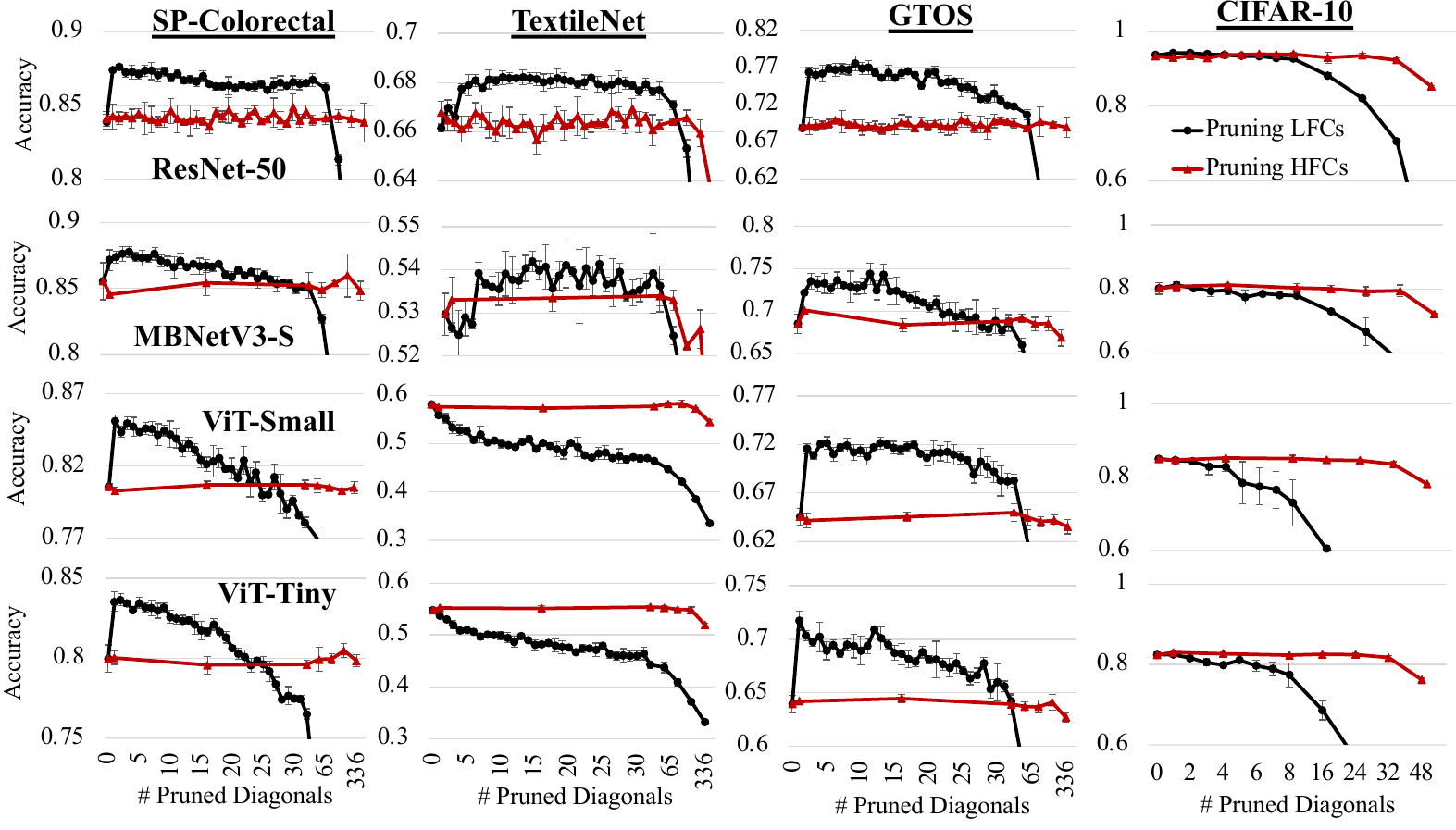}
   \caption{ID Results for ResNet-50, MobileNet-V3, ViT-Small, and ViT-Tony.}
   \label{fig:id_all_models}
\end{figure}

\begin{figure}[!ht]
  \centering
   \includegraphics[width=1\linewidth]{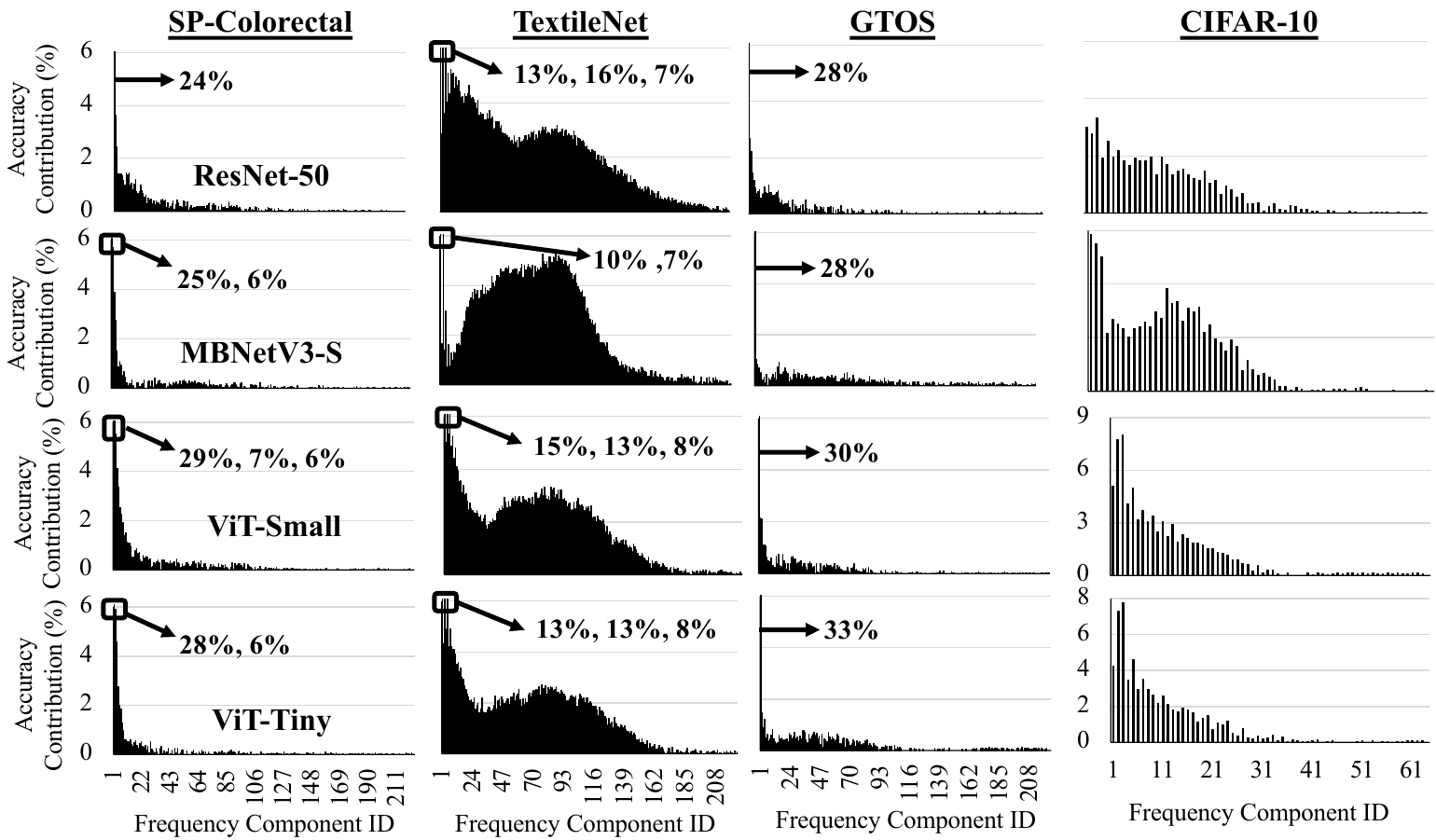}
   \caption{Spectral Behavior for ResNet-50, MobileNet-V3, ViT-Small, and ViT-Tony when trained with unpruned images.}
   \label{fig:spec_behv_unpruned_all_models}
\end{figure}

\begin{figure}[!ht]
  \centering
   \includegraphics[width=1\linewidth]{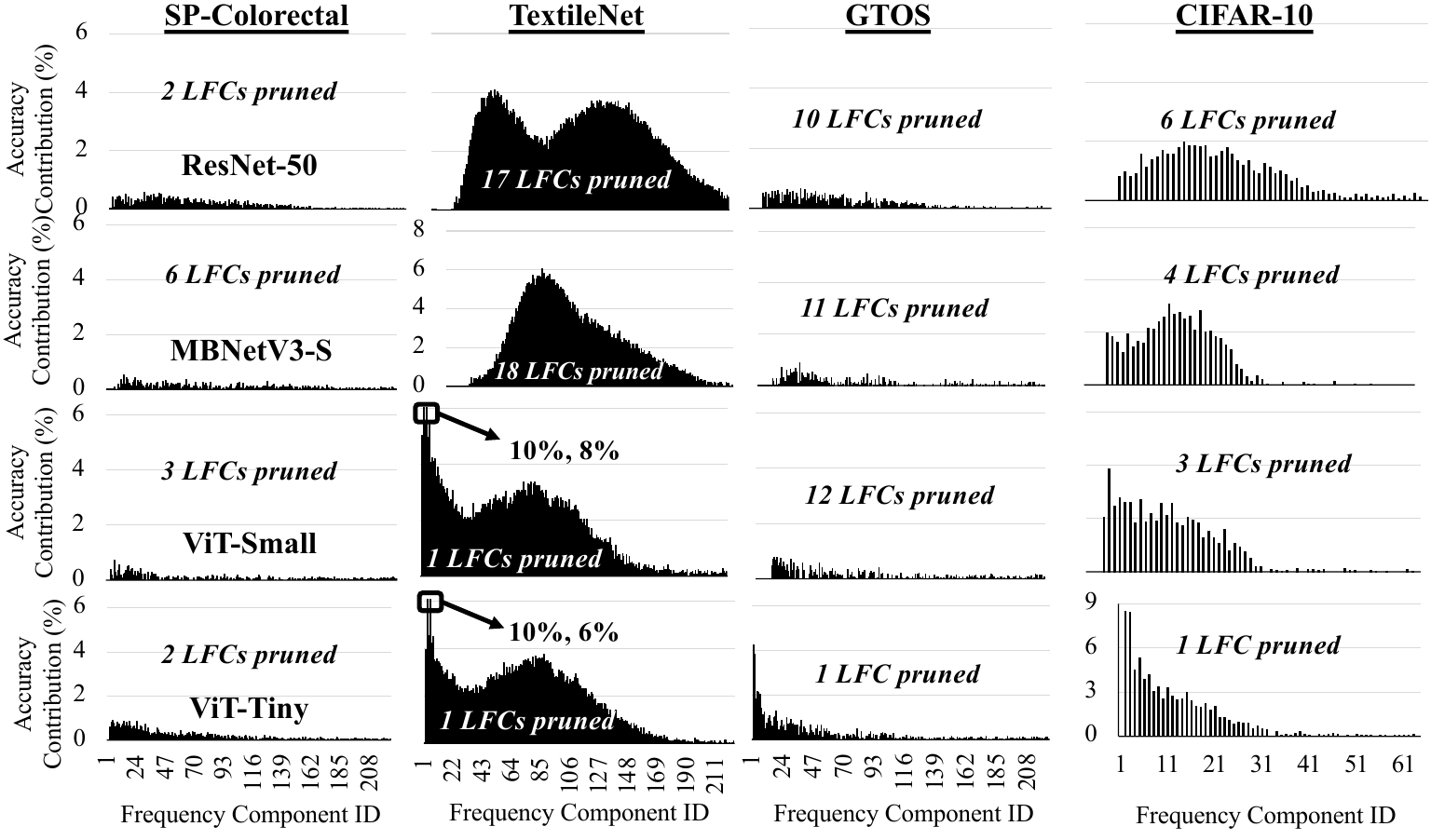}
   \caption{Spectral Behavior for ResNet-50, MobileNet-V3, ViT-Small, and ViT-Tony when trained with pruned images.}
   \label{fig:spec_behv_pruned_all_models}
\end{figure}

\section{Impact of Model Size and Architecture on ID Results}
\label{sect:app:id_spec_all_models}

Figure \ref{fig:id_all_models} presents ID accuracy results for pruning LFCs and HFCs for all the four neural architectures we analyzed: ResNet-50, MobileNet-V3, ViT-Small, and ViT-Tiny. Figure \ref{fig:spec_behv_unpruned_all_models} and \ref{fig:spec_behv_pruned_all_models} show the spectral behavior for the four models we use. As can be seen, the results are consistent across different neural architectures; pruning HFCs do not significantly impact the ID accuracy, whereas pruning LFCs improves it for all texture-driven tasks, except TextileNet with ViT-Small and ViT-Tiny. Nevertheless, all the models and datasets have a skewed behavior towards the lowest-frequency coefficients.

Decreasing ID results for TextileNet with ViTs is due to two reasons: (i) ViTs have a stronger bias towards LFCs than CNNs, and (ii) TextileNet has a stronger low-frequency noise than SP-Colorectal and GTOS.

\begin{wrapfigure}{r}{0.55\linewidth}
  \centering
  \vspace{-0.8\baselineskip}
  \includegraphics[width=\linewidth]{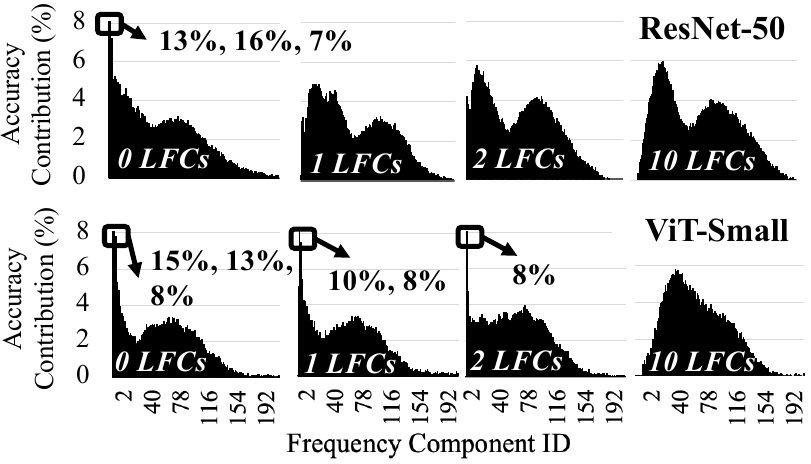}
  \caption{Spectral behavior of TextileNet.}
  \label{fig:txt_r50_vit}
  \vspace{-0.8\baselineskip}
\end{wrapfigure}

Figure \ref{fig:txt_r50_vit} presents accuracy contributions for TextileNet with ResNet-50 (top) and ViT-Small (bottom) for an increasing number of pruned LFCs. As shown, pruning LFCs does not immediately eliminate the skewed distribution, i.e., the low-frequency shortcut; whereas, for ResNet-50, where pruning even a single LFC immediately eliminates the skewed distribution and shifts the distribution towards HFCs. CNNs use fine-grained strided convolutions over the whole image, which allow them to more easily extract fine-grained, high-frequency information than ViTs, which use disjoint image patches and hence have a stronger bias towards low-frequency noises. There is a recent line of work reporting similar results on ViTs versus CNNs \cite{bai2022,pan2022,park2022,rao2021,wang2022,wang_zhenyu_2022}.

Secondly, TextileNet images are fashion models wearing garments made from different fabrics (see Figure 3 of our submission). These images contain more severe low-frequency noise, compared to SP-Colorectal and GTOS, e.g., in the photographic background and the body shape. SP-Colorectal and GTOS, on the other hand, have squared images of tissue scans and ground terrains. As a result, ViTs suffer more severely from low-frequency shortcuts for TextileNet than they do for SP-Colorectal and GTOS.

Therefore, low-frequency shortcuts are universal phenomena across different model sizes and architectures for texture-driven tasks. Pruning, however, is not a universally successful method to eliminate the low-frequency shortcut. In particular, when the model architecture has a strong bias towards low-frequency information, such as ViTs, and also the domain includes heavy low-frequency noise, e.g., TextileNet, pruning falls short on eliminating the shortcut. Nevertheless, in 10 of 12 cases (4 models \& 3 datasets), pruning LFCs improves the ID performance.

\begin{figure*}[!ht]
  \centering
   \includegraphics[width=1\linewidth]{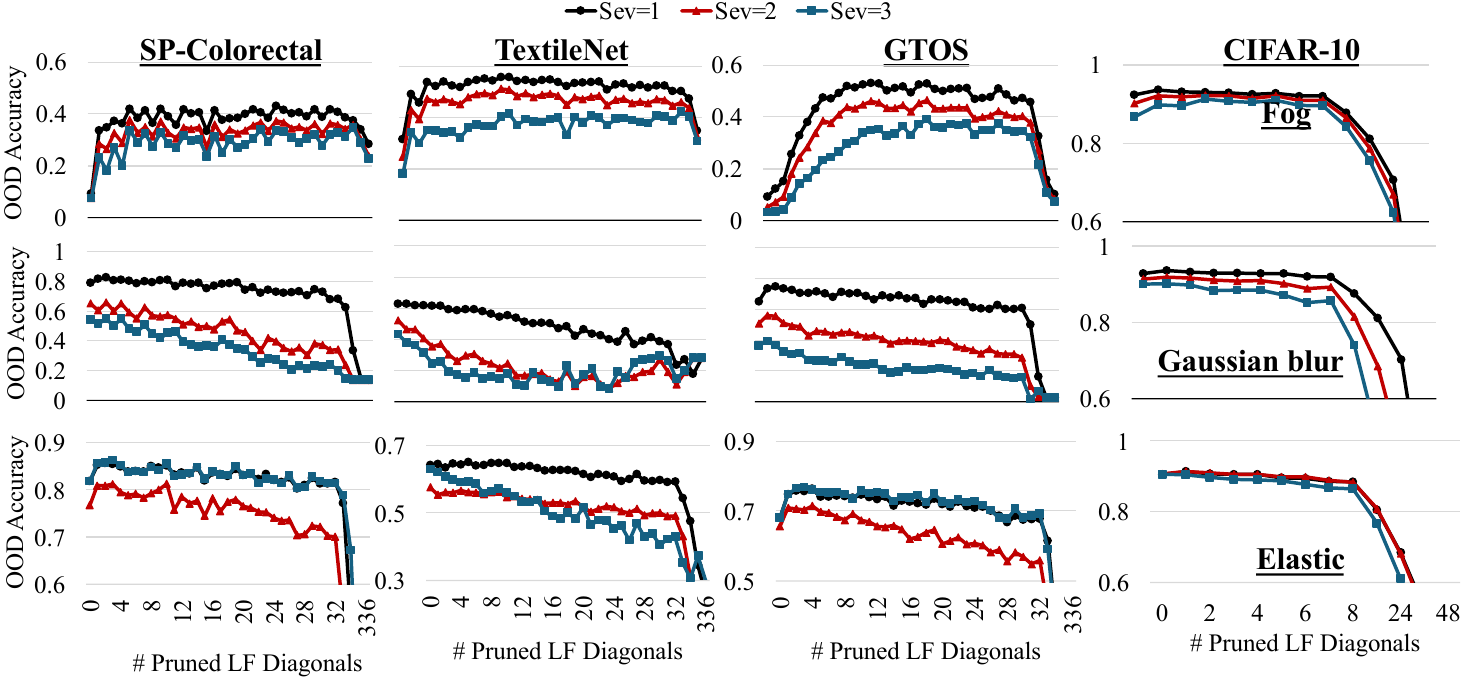}
   \caption{OOD performance under different severity levels. X-axis: number of pruned LFCs. Y-axis: OOD accuracy. Overall patterns remain similar across different severity levels. Fog's OOD accuracy improves as the low-frequency shortcut is eliminated. Gaussian blur and elastic face a trade-off between the improved representation and magnified corruptions at the HFCs.}
   \label{fig:ood_sev}
\end{figure*}

\section{Results Across OOD Severity Levels}
\label{sect:app:ood_severity}

Figure \ref{fig:ood_sev} presents OOD results for the four tasks we study across three severity levels: 1, 2, and 3. The x-axis shows the number of pruned LF diagonals, whereas the y-axis shows the OOD accuracy. As can be seen, the main patterns remain the same across the severity levels. Texture-driven tasks have improved accuracy for fog corruption, as fog is low-frequency-heavy, and texture-driven tasks suffer from low-frequency shortcuts. Gaussian blur and elastic transform can increase, stabilize, or decrease accuracy, depending on the trade-off between improved representation and magnified high-frequency corruption. CIFAR-10 has a small accuracy drop at the severity of 1, whereas a more visible accuracy degradation at the severity of 2 and 3.

\begin{figure}[!ht]
  \centering
   \includegraphics[width=1\linewidth]{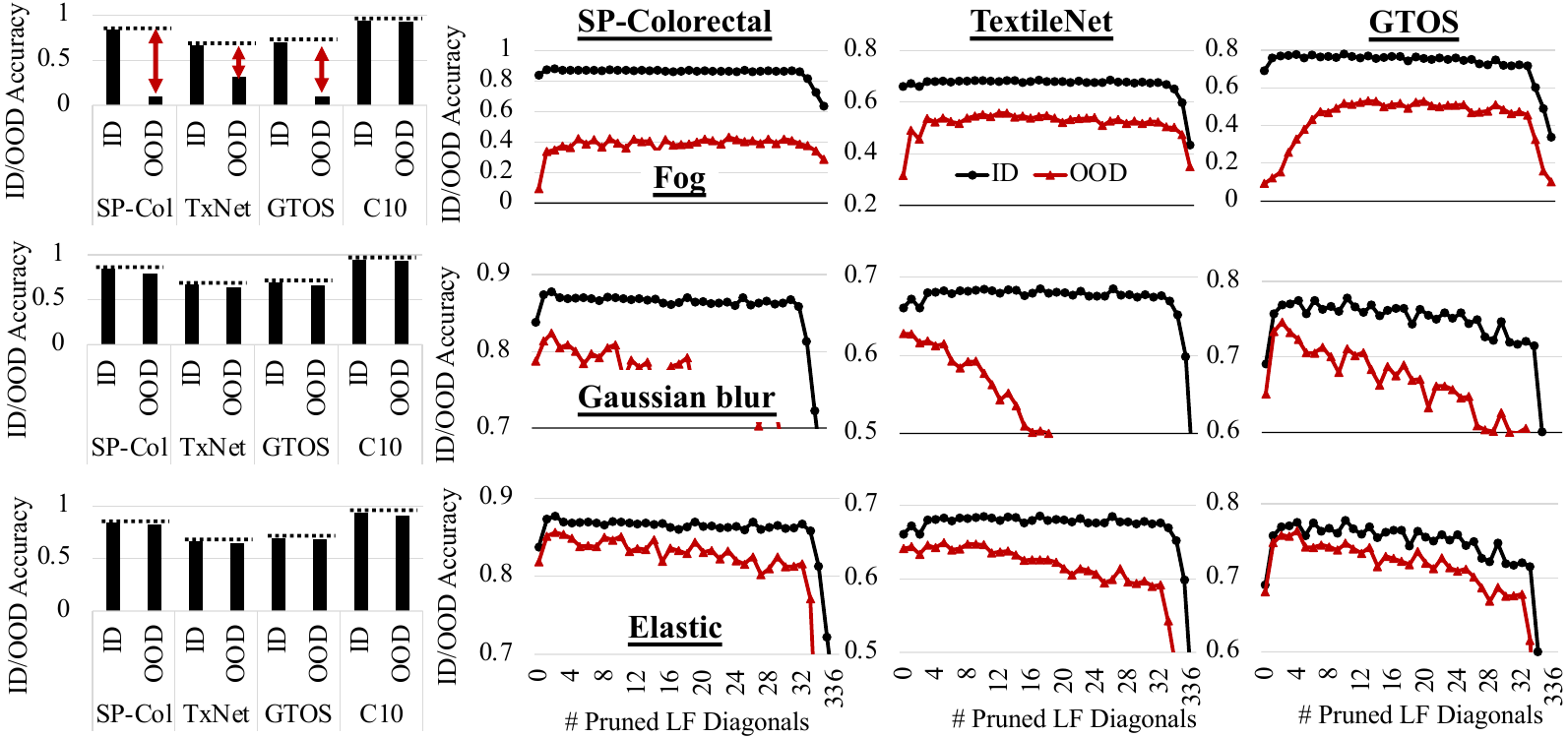}
   \caption{OOD Results for ResNet50.}
   \label{fig:ood_resnet50_all}
\end{figure}

\begin{figure}[!ht]
  \centering
   \includegraphics[width=1\linewidth]{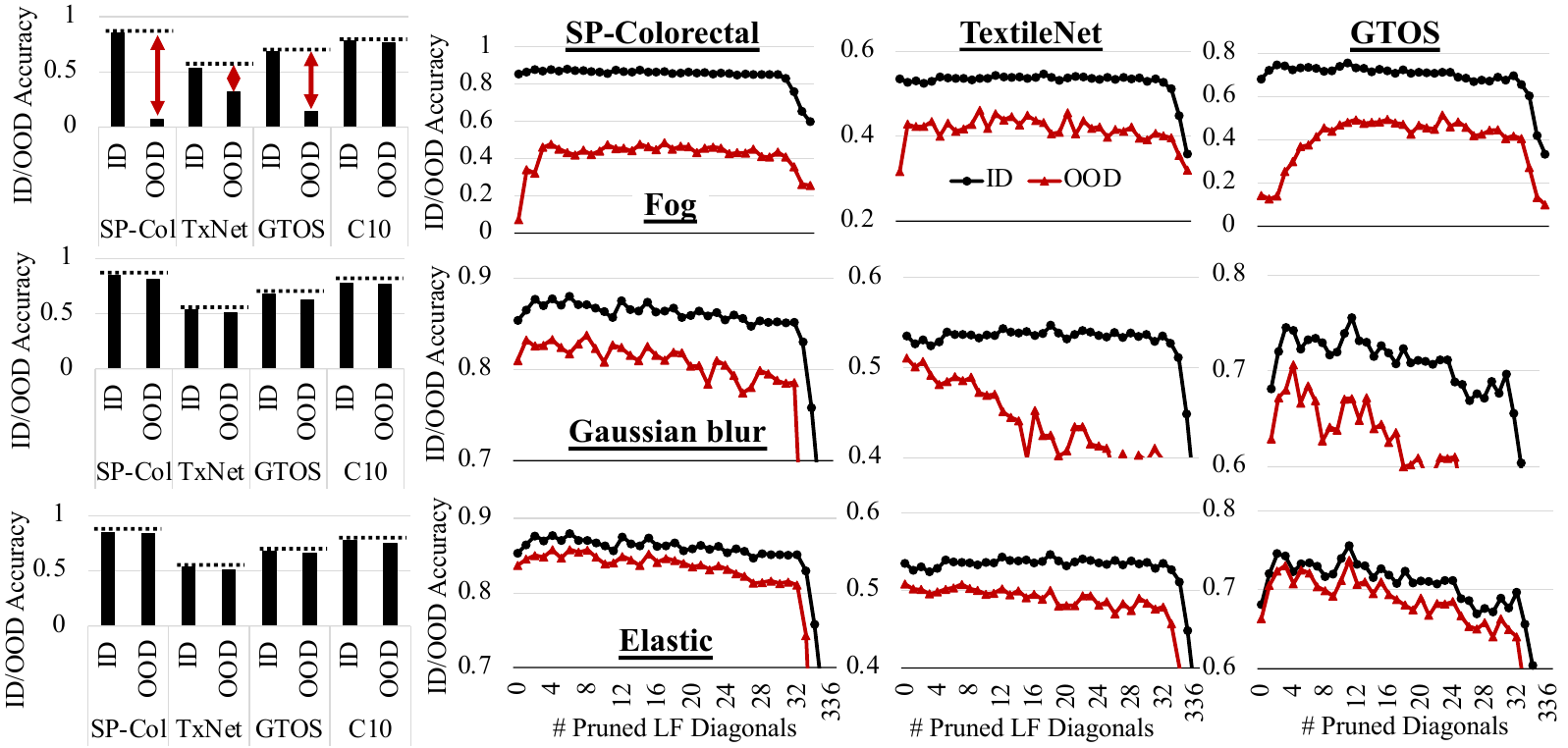}
   \caption{OOD Results for MobileNet-V3.}
   \label{fig:ood_mbnet_all}
\end{figure}

\begin{figure}[!ht]
  \centering
   \includegraphics[width=1\linewidth]{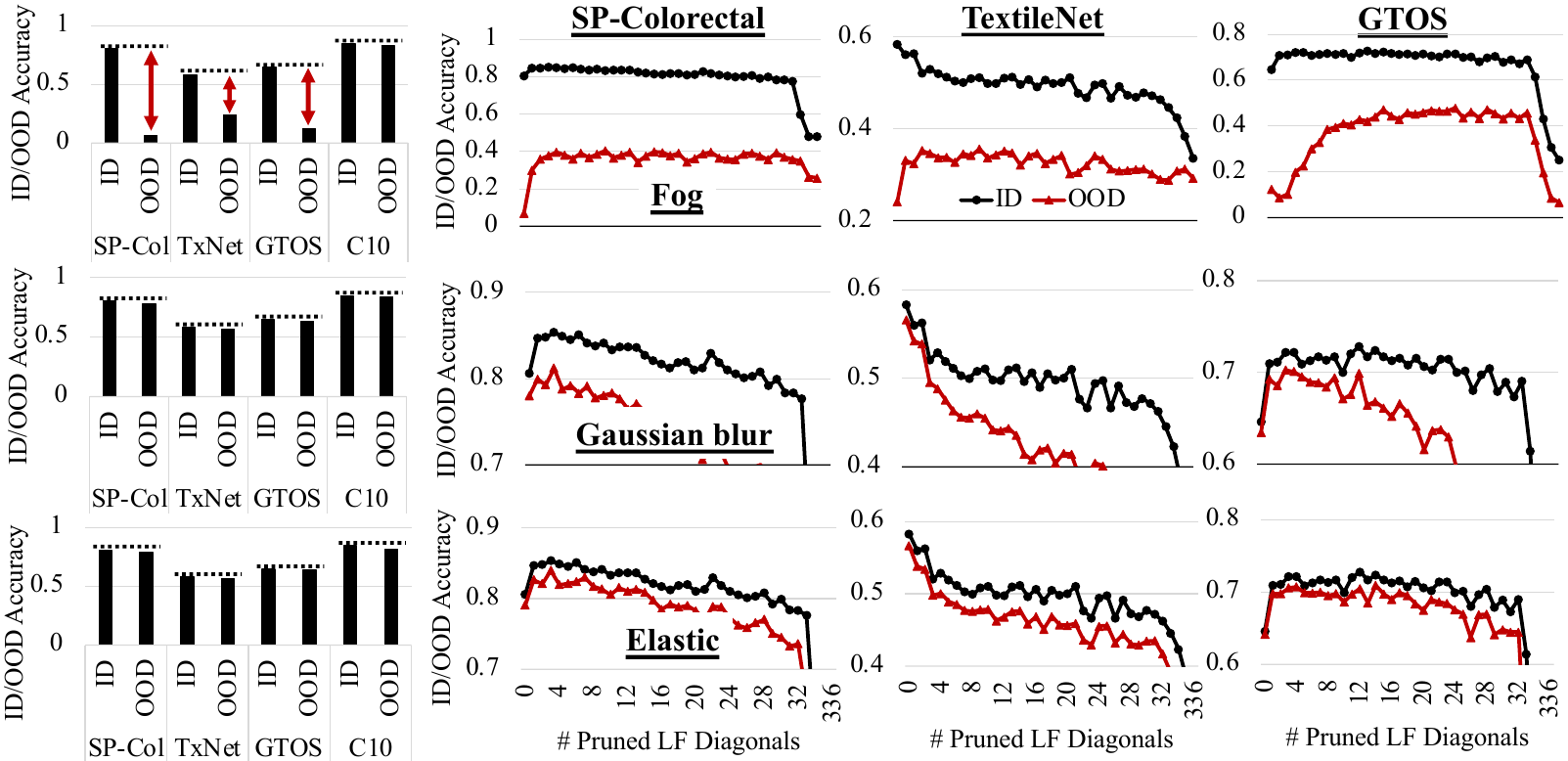}
   \caption{OOD Results for ViT-Small.}
   \label{fig:ood_vit_small_all}
\end{figure}

\begin{figure}[!ht]
  \centering
   \includegraphics[width=1\linewidth]{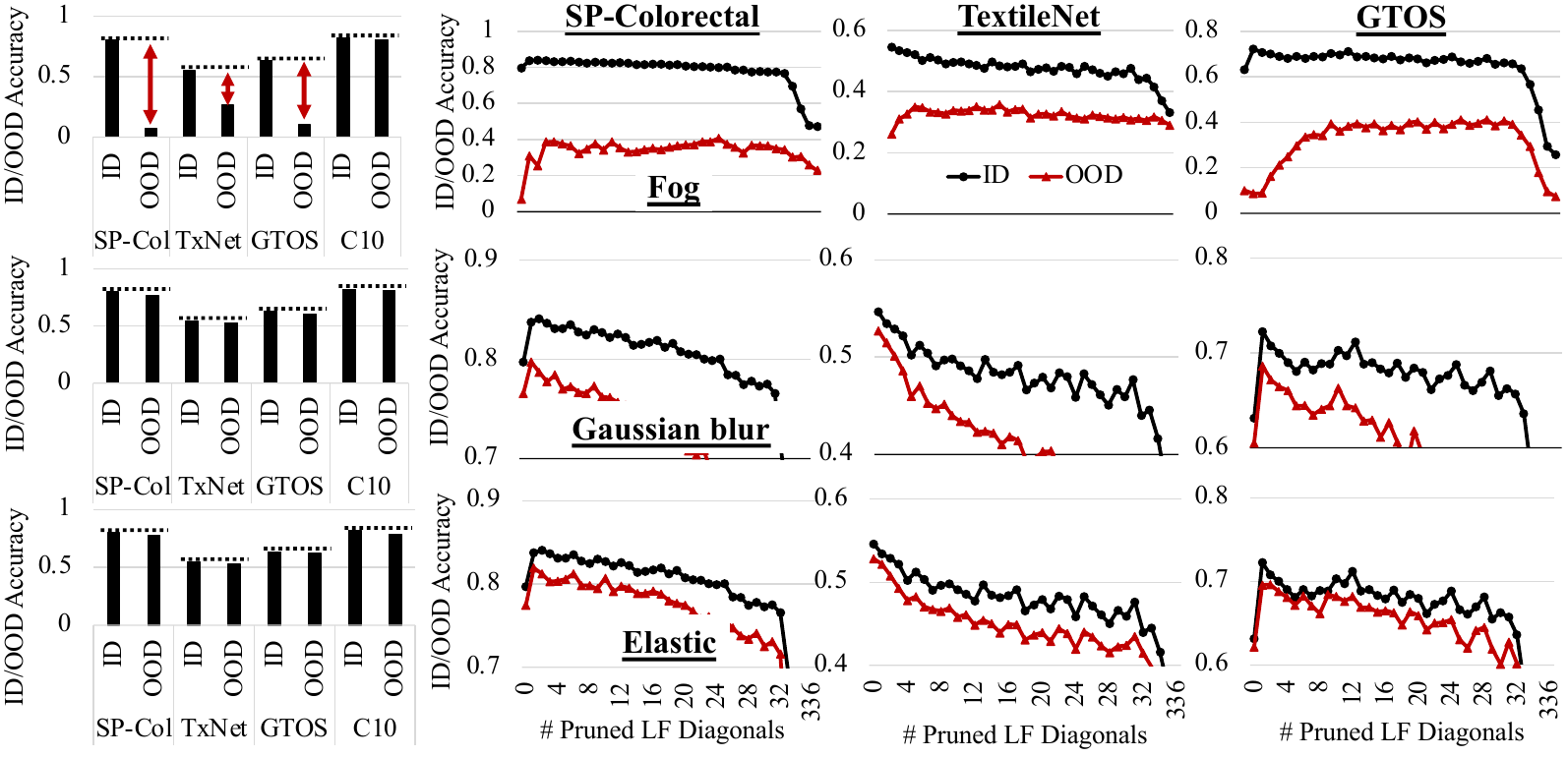}
   \caption{OOD Results for ViT-Tiny.}
   \label{fig:ood_vit_tiny_all}
\end{figure}

\section{Impact of Model Size and Architecture on OOD Results}
\label{sect:app:ood_all_models}

Figure \ref{fig:ood_resnet50_all} to Figure \ref{fig:ood_vit_tiny_all} presents the OOD results for ResNet-50, MobileNet-V3, ViT-Small, and ViT-Tiny for the three corruption types we analyze, fog (low-frequency corruption), Gaussian blur (high-frequency corruption), and mixed low-and-high-frequeny corruption (elastic transform). As can be seen, trends are similar across different model sizes and architectures. For each model, we report the results for a single training with seed 42. Earlier in the appendix, we have shown that different seeds produce similar OOD results (see Appendix \ref{sect:app:seeds}).

\begin{figure*}[!ht]
  \centering
   \includegraphics[width=.9\linewidth]{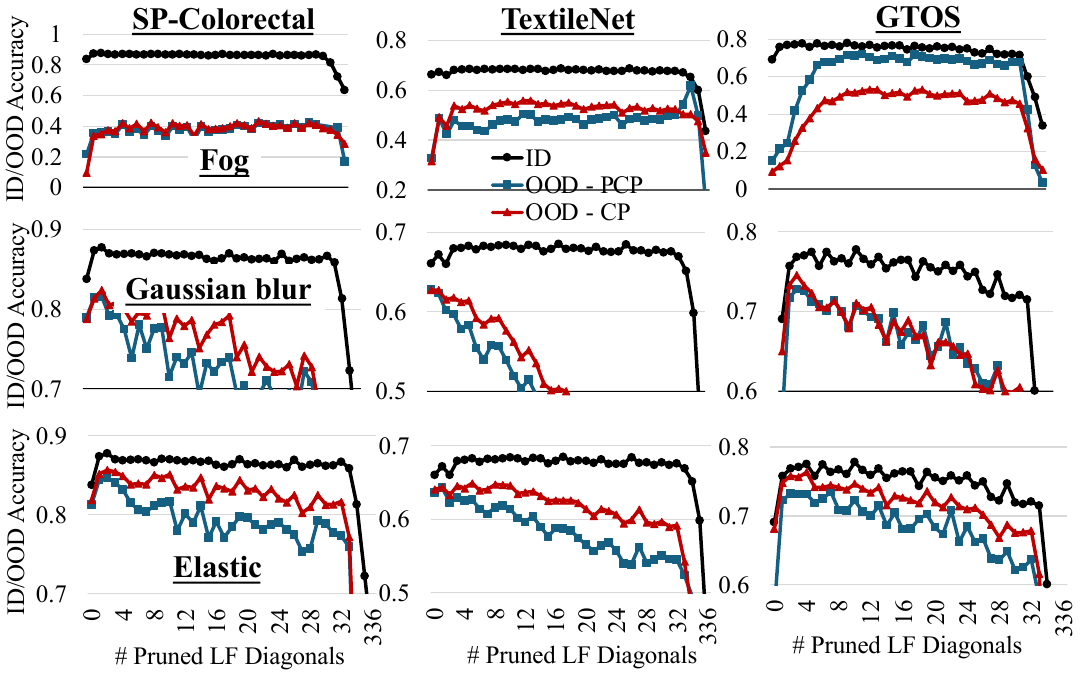}
   \caption{OOD corruption pipelines. PCP: prune-corrupt-prune. CP: corrupt-prune. \textbf{Top:} Low-frequency corruption (fog) results. \textbf{Middle:} High-frequency corruption (Gaussian blur) results. \textbf{Bottom:} Mixed-frequency corruption (elastic transform) results. PCP and CP follow similar trends across the tasks and corruption types.}
   \label{fig:ood_accuracy_pipelines}
\end{figure*}

\section{OOD Corruption Pipelines}
\label{sect:app:ood_pipeline}

Figure \ref{fig:ood_accuracy_pipelines} presents the OOD results for the two corruption pipelines we tested: (i) corrupt-prune (CP), and (ii) prune-corrupt-prune (PCP). In Section \ref{sect:ood}, we covered the first pipeline. In this section, we present the results for both pipelines. The x-axis presents the number of pruned LFCs, and the y-axis presents the ID/OOD accuracy. While the red line shows the CP pipeline that we also present in Section \ref{sect:ood}, the blue line shows the PCP pipeline. Dark line presents the ID values for reference. As can be seen, blue and red lines are often close to each other and follow similar patterns. For the GTOS task and fog corruption, the PCP pipeline has a higher accuracy than the CP pipeline. In fact, the PCP pipeline largely recovers from the corruptions, closely approximating the ID accuracy. This is because applying corruption after pruning \textit{reduces} the impact of corruption, as corruption uses some of the pruned components. As a result, the final OOD accuracy is higher for PCP than for CP.

For elastic corruption, however, CP achieves higher accuracy than PCP. This is because applying corruption after pruning \textit{increases} its impact. Elastic transform spreads local pixels to their neighbors, which worsens its effect in the absence of LFCs. Which pipeline causes more harm depends on the exact formula the corruption uses and the dataset's characteristics. 

Nevertheless, OOD accuracy follows similar trends compared to the ID accuracy for both pipelines. While LFC-pruning significantly helps recover from fog corruption, it introduces a trade-off for Gaussian blur and elastic transform corruptions. On the one hand, pruning LFCs improves representation, as HFCs better reflect the application semantics. On the other hand, the model's focus shifts to HFCs once LFCs are pruned. Hence, corruption at the HFCs is magnified. OOD accuracy increases/decreases/remains stable, depending on this trade-off.

\begin{figure*}[!ht]
  \centering
   \includegraphics[width=0.9\linewidth]{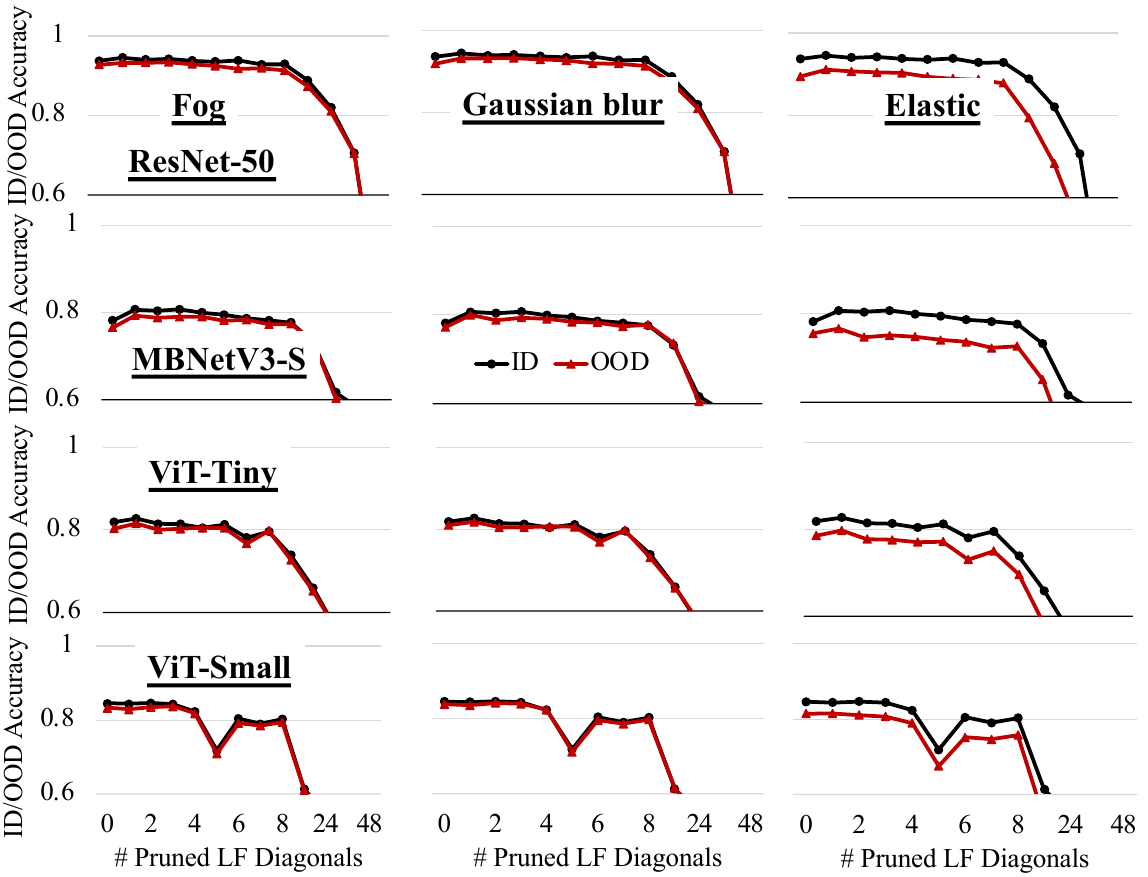}
   \caption{CIFAR-10's OOD performance closely approximates its ID performance. It suffers an observable drop in elastic corruption due to the elastic transform’s broad coverage of corruption across low and high frequencies.}
   \label{fig:ood_cifar10}
\end{figure*}

\section{OOD Results on CIFAR-10}
\label{sect:app:cifar10_ood}

CIFAR-10's OOD values closely approximate its ID values, thanks to its balanced spectral behavior shown in Figure \ref{fig:spec_behv_resnet50}. Figure \ref{fig:ood_cifar10} presents ID and OOD values for CIFAR-10 for the three corruptions we analyze. As shown, OOD values (red line) closely approximate the ID values (dark line). For elastic corruption, OOD values are observed to be lower. This is because elastic modifies a wide range of frequency components, including both low and high frequencies, which reduces the models' tolerance capacity.

\begin{figure}[!h]
  \centering
   \includegraphics[width=1\linewidth]{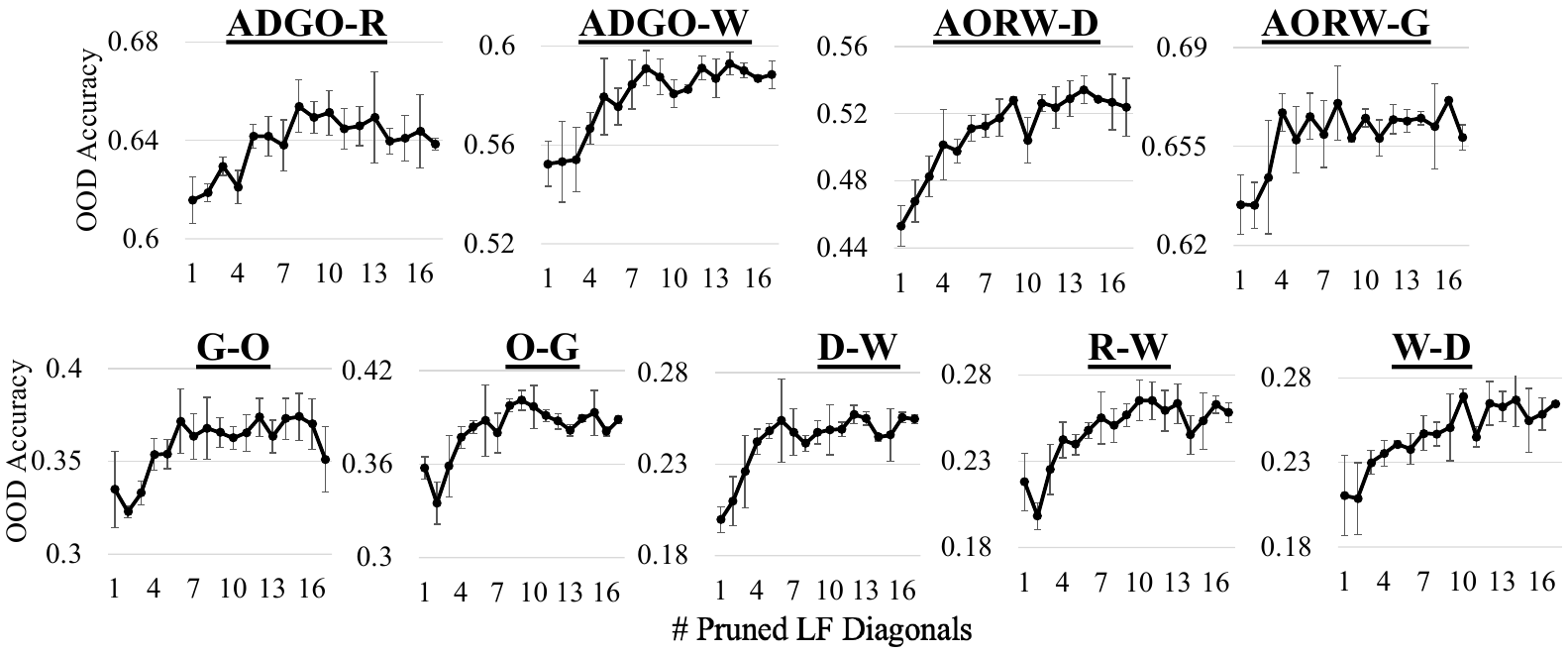}
   \caption{OOD results for real-world domain-generalization. Each graph represents results for a different subset of NICO++.}
   \label{fig:nicopp_ood_all}
\end{figure}

\begin{figure}[!h]
  \centering
   \includegraphics[width=.9\linewidth]{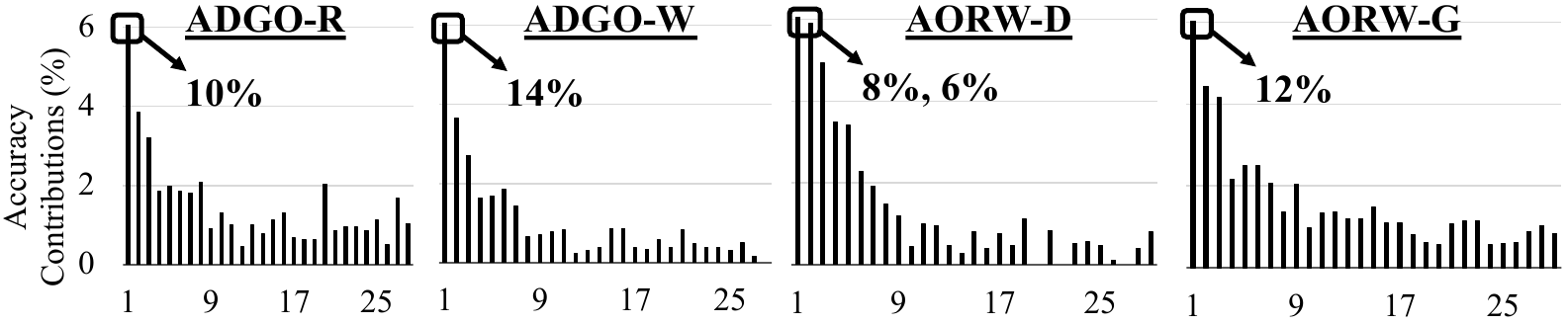}
   \caption{Spectral behavior for real-world domain-generalization. Each graph represents results for a different subset of NICO++.}
   \label{fig:nicopp_spec_behv_all}
\end{figure}

\section{Results on Real-World Domain-Generalization}
\label{sect:app:nicopp_all}

Figure \ref{fig:nicopp_ood_all} presents OOD results for NICO++ for nine different combinations of training and validation sets using ResNet-50. As can be seen, pruning LFCs consistently improves accuracy. Figure \ref{fig:nicopp_spec_behv_all} presents the spectral behavior for the four subsets, ADGO-R, ADGO-W, AORW-D, and AORW-D. As can be seen, spectral behavior is always skewed towards LFCs, even though being at different levels depending on the specific training/validation set combination. These results show that low-frequency shortcuts play an important role for domain-generalization problems. Neural nets are highly biased towards learning from background cues. Pruning LFCs forces the network to learn from more realistic higher-frequency features and hence achieve a better OOD generalization performance.

\begin{figure}[!h]
  \centering
   \includegraphics[width=0.73\linewidth]{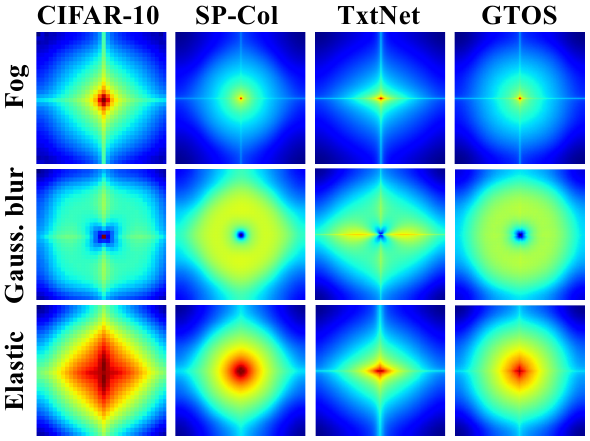}
   \caption{Frequency characteristics of the corruptions we use, across the four tasks we analyze. We use a log10-scale, unlike \cite{yin2019} in their Figure 2. This way, we can observe fog's high-frequency corruptions, which would otherwise go unnoticed. Each image represents the Fourier characteristics of the average difference between the original and corrupted image. We shift the lowest frequency coefficient to the center for compatibility with \cite{yin2019}.}
   \label{fig:ood_freq_char}
\end{figure}

\begin{table}[t]
\centering
\caption{Hyperparameter search space for ResNet-50.}
\label{tab:resnet_hyperparameter_search_space}
\small
{%
\begin{tabular}{llccccc}
\toprule
HID & Optimizer & Learning rate & Momentum & Weight decay & Scheduler & Epochs \\
\midrule
0 & SGD  & $3 \times 10^{-2}$ & 0.9 & $5 \times 10^{-4}$ & Cosine    & 200 \\
1 & SGD  & $1 \times 10^{-1}$ & 0.9 & $5 \times 10^{-4}$ & Cosine    & 200 \\
2 & SGD  & $2 \times 10^{-1}$ & 0.9 & $5 \times 10^{-4}$ & Cosine    & 200 \\
3 & SGD  & $1 \times 10^{-1}$ & 0.9 & $1 \times 10^{-4}$ & Cosine    & 200 \\
4 & SGD  & $1 \times 10^{-1}$ & 0.9 & $1 \times 10^{-3}$ & Cosine    & 200 \\
5 & SGD  & $1 \times 10^{-1}$ & 0.9 & $5 \times 10^{-4}$ & Multistep & 200 \\
0 & Adam & $1 \times 10^{-4}$ & --  & $1 \times 10^{-4}$ & Cosine    & 200 \\
1 & Adam & $3 \times 10^{-4}$ & --  & $1 \times 10^{-4}$ & Cosine    & 200 \\
2 & Adam & $1 \times 10^{-3}$ & --  & $1 \times 10^{-4}$ & Cosine    & 200 \\
3 & Adam & $3 \times 10^{-4}$ & --  & $0$                 & Cosine    & 200 \\
4 & Adam & $3 \times 10^{-4}$ & --  & $5 \times 10^{-4}$ & Cosine    & 200 \\
5 & Adam & $3 \times 10^{-4}$ & --  & $1 \times 10^{-4}$ & Multistep & 200 \\
\bottomrule
\end{tabular}%
}
\end{table}

\begin{table}[t]
\centering
\caption{Hyperparameter search space for ViT-Small. No scheduler or momentum used.}
\label{tab:vit_hyperparameter_search_space}
\small
\begin{tabular}{llcccc}
\toprule
HID & Optimizer & Learning rate & Weight decay & Warmup epochs & Epochs \\
\midrule
0 & AdamW & $1 \times 10^{-3}$ & 0.05 & 3.4 & 300 \\
1 & AdamW & $3 \times 10^{-3}$ & 0.05 & 3.4 & 300 \\
2 & AdamW & $5 \times 10^{-3}$ & 0.05 & 3.4 & 300 \\
3 & AdamW & $3 \times 10^{-3}$ & 0.10 & 3.4 & 300 \\
4 & AdamW & $3 \times 10^{-3}$ & 0.30 & 3.4 & 300 \\
5 & AdamW & $3 \times 10^{-3}$ & 2.50 & 3.4 & 300 \\
\bottomrule
\end{tabular}
\end{table}

\begin{table}[t]
\centering
\caption{GTOS ID performance with ResNet-50 using SGD and Adam across six hyperparameters each, as we prune increasing number of LFCs. As can be seen, pruning LFCs increases ID accuracy for both optimizers and all hyperparameters.}
\label{tab:gtos_resnet50_sgd_adam}
\resizebox{\textwidth}{!}{%
\begin{tabular}{lcccccc|cccccc|cc}
\toprule
& \multicolumn{6}{c|}{GTOS w/ ResNet-50 \& SGD}
& \multicolumn{6}{c|}{GTOS w/ ResNet-50 \& Adam}
& & \\
\cmidrule(lr){2-7}\cmidrule(lr){8-13}
& hid=0 & hid=1 & hid=2 & hid=3 & hid=4 & hid=5
& hid=6 & hid=7 & hid=8 & hid=9 & hid=10 & hid=11
& avg. SGD & avg. Adam \\
\midrule
0-LFC & 0.706557 & 0.693443 & 0.698033 & 0.665574 & 0.699016 & 0.680328 & 0.674426 & 0.712787 & 0.680984 & 0.698689 & 0.705574 & 0.710820 & 0.690492 & 0.697213 \\
1-LFC & 0.775410 & 0.765574 & 0.760656 & 0.738033 & 0.759016 & 0.763934 & 0.746557 & 0.780000 & 0.762295 & 0.747213 & 0.755738 & 0.771475 & 0.760437 & 0.760546 \\
2-LFC & 0.761311 & 0.755738 & 0.746230 & 0.735410 & 0.761311 & 0.764262 & 0.745574 & 0.763934 & 0.758689 & 0.752787 & 0.767213 & 0.767213 & 0.754044 & 0.759235 \\
3-LFC & 0.775410 & 0.770164 & 0.765902 & 0.729180 & 0.778033 & 0.748852 & 0.752459 & 0.777377 & 0.760328 & 0.766885 & 0.761967 & 0.771475 & 0.761257 & 0.765082 \\
4-LFC & 0.789180 & 0.773115 & 0.765246 & 0.742295 & 0.767541 & 0.769836 & 0.748852 & 0.767869 & 0.761311 & 0.757705 & 0.777049 & 0.770492 & 0.767869 & 0.763880 \\
5-LFC & 0.780656 & 0.769508 & 0.768197 & 0.726557 & 0.760000 & 0.757049 & 0.744262 & 0.771475 & 0.748525 & 0.754426 & 0.760656 & 0.769508 & 0.760328 & 0.758142 \\
6-LFC & 0.766230 & 0.769180 & 0.774754 & 0.724590 & 0.767213 & 0.765574 & 0.743607 & 0.763934 & 0.744918 & 0.745902 & 0.773115 & 0.776393 & 0.761257 & 0.757978 \\
7-LFC & 0.761967 & 0.760984 & 0.760328 & 0.729836 & 0.783279 & 0.760000 & 0.738689 & 0.779016 & 0.749836 & 0.768197 & 0.770164 & 0.758033 & 0.759399 & 0.760656 \\
8-LFC & 0.774754 & 0.768525 & 0.762951 & 0.720328 & 0.768525 & 0.757705 & 0.740984 & 0.757377 & 0.751148 & 0.747213 & 0.769836 & 0.753115 & 0.758798 & 0.753279 \\
\bottomrule
\end{tabular}%
}
\end{table}

\begin{table}[t]
\centering
\caption{SP-Colorectal and GTOS ID performance with ViT-Small using Adam across six hyperparameters each, as we prune increasing number of LFCs. As can be seen, pruning LFCs increases ID accuracy for both datasets and all hyperparameters.}
\label{tab:vit_small_adam_sp_colorectal_gtos}
\resizebox{\textwidth}{!}{%
\begin{tabular}{lcccccc|c|cccccc|c}
\toprule
& \multicolumn{7}{c|}{SP-Colorectal w/ ViT-Small \& Adam}
& \multicolumn{7}{c}{GTOS w/ ViT-Small \& Adam} \\
\cmidrule(lr){2-8}\cmidrule(lr){9-15}
& hid=0 & hid=1 & hid=2 & hid=3 & hid=4 & hid=5 & avg.
& hid=0 & hid=1 & hid=2 & hid=3 & hid=4 & hid=5 & avg. \\
\midrule
0-LFC & 0.791008 & 0.783392 & 0.786028 & 0.802871 & 0.803017 & 0.743117 & 0.784906 & 0.624918 & 0.345574 & 0.596721 & 0.602623 & 0.643607 & 0.634426 & 0.574645 \\
1-LFC & 0.826889 & 0.830258 & 0.819127 & 0.838459 & 0.842267 & 0.769479 & 0.821080 & 0.705574 & 0.367541 & 0.537705 & 0.651475 & 0.718033 & 0.659344 & 0.606612 \\
2-LFC & 0.825425 & 0.829379 & 0.827914 & 0.846514 & 0.842414 & 0.755419 & 0.821178 & 0.699344 & 0.423934 & 0.455410 & 0.661639 & 0.700984 & 0.670492 & 0.601967 \\
3-LFC & 0.830844 & 0.833333 & 0.835970 & 0.847832 & 0.839192 & 0.742824 & 0.821666 & 0.698361 & 0.474426 & 0.347213 & 0.681967 & 0.718361 & 0.635738 & 0.592678 \\
4-LFC & 0.828500 & 0.827182 & 0.829233 & 0.837141 & 0.845489 & 0.758055 & 0.820933 & 0.693115 & 0.400656 & 0.530820 & 0.675738 & 0.733115 & 0.631475 & 0.610820 \\
5-LFC & 0.836262 & 0.832162 & 0.839484 & 0.849151 & 0.842999 & 0.753661 & 0.825620 & 0.709180 & 0.363934 & 0.583934 & 0.685574 & 0.718033 & 0.617377 & 0.613005 \\
6-LFC & 0.832601 & 0.825864 & 0.833919 & 0.839777 & 0.848858 & 0.749268 & 0.821715 & 0.695410 & 0.413443 & 0.517049 & 0.682951 & 0.722951 & 0.623607 & 0.609235 \\
7-LFC & 0.839484 & 0.832748 & 0.842560 & 0.841681 & 0.846661 & 0.756444 & 0.826596 & 0.700000 & 0.393443 & 0.624918 & 0.684918 & 0.708525 & 0.614754 & 0.621093 \\
8-LFC & 0.832601 & 0.832748 & 0.838899 & 0.838606 & 0.846514 & 0.740920 & 0.821715 & 0.693443 & 0.368197 & 0.613443 & 0.689836 & 0.719672 & 0.616066 & 0.616776 \\
\bottomrule
\end{tabular}%
}
\end{table}

\section{Tuning Hyperparameters for Low-Frequency Shortcuts}
\label{sect:app:hypparam}

We test GTOS and SP-Colorectal across different optimizers, SGD and Adam, using six different hyperparameters for ResNet-50 and ViT-Small. We list the hyperparameter search space for ResNet-50 in Table \ref{tab:resnet_hyperparameter_search_space} and for ViT-Small in \ref{tab:vit_hyperparameter_search_space}. Table \ref{tab:gtos_resnet50_sgd_adam} presents the ID results for ResNet-50 hyperparameter search space. As can be seen, pruning LFCs increases the ID accuracy for all the hyperparameters for both optimizers.  Table \ref{tab:vit_small_adam_sp_colorectal_gtos} presents the ID results for ViT-Small hyperparameters. Once again, we observe that pruning LFCs improves the ID accuracy for all the hyperparameters for both SP-Colorectal and GTOS datasets. This shows that low-frequency shortcuts are persistent across different hyperparameters.

\section{Additional Comparison to Related Work}
\label{sect:app:relwork}

\noindent \textbf{Existing CIFAR-10 Studies.} There are three main frequency characterization studies on CIFAR-10. \cite{jo2017} performs test set characterization on unpruned training images and concludes that high frequencies are important for CIFAR-10 classification. They prune all HFCs with an above radial frequency of 4.25 from test set images and observe up to 24\% accuracy drop in test set accuracy\footnote{See Figure 8 of \cite{jo2017}, a \& b. First row with the first and second columns.}. Our results corroborate this, as the radial frequency of 4.25 corresponds to the top-left 8 LFCs in our setup. Pruning everything beyond 8 LFCs can indeed cause a significant accuracy drop, as our accuracy contributions graph also shows in \ref{fig:spec_behv_resnet50}. \cite{abello2021} also performs test set characterization with a fixed unpruned training set of images, and show that pruning the highest end of the frequency spectrum causes the least accuracy loss\footnote{See Figure 5 of \cite{abello2021}, top row, (15-32) data point on the x-axis.}, as we also have shown in our experiments. They further show that pruning higher frequencies in the lower half of the spectrum can cause a larger drop in accuracy than pruning lower frequencies. Our results corroborate this, as we have shown that the lowest 22 components in the lower half of the spectrum contribute roughly equally to test set accuracy. \cite{wang2020} performs training and test set characterization with a small number of configurations and shows that keeping LFCs contributes significantly more than HFCs to the test accuracy\footnote{See Table 1 of \cite{wang2020}.}, which corroborates our results.

\noindent \textbf{Existing ImageNet Studies.} Our CIFAR-10 results on LFC and HFC pruning from training and test sets corroborate with numerous existing ImageNet \cite{imagenet} results. Task-specific image compression studies by \cite{sirin2024} and \cite{xu_cvpr2020} show that pruning HFCs is much more advantageous than LFCs for a subset and full ImageNet\footnote{See Figure 5 of \cite{sirin2024}, and Figure 5 of \cite{xu_cvpr2020}.}. Similarly, \cite{yin2019} shows that accuracy drops much faster when training and test set images are high-pass filtered, as opposed to low-pass filtered\footnote{See Figure 1 of \cite{yin2019}.}. Lastly, \cite{abello2021} shows a similar trend to ours for a subset of ImageNet, as they prune high frequencies\footnote{See Figure 5 of \cite{abello2021}, third figure from the top.}. 
These studies, however, did not conduct as fine-grained an analysis as we have. Their unit of analysis is more than a single diagonal. Hence, their results might be incomplete. Further analysis of ImageNet and similar datasets is in our future work.

Existing studies either group frequency coefficients in broader units than diagonals \cite{abello2021,jo2017}, or perform an analysis one by one for each frequency coefficient separately \cite{shortcut1}. Using broad units limits the findings of the existing studies. Performing an analysis for each coefficient is feasible for a test set characterization, but it is too costly for a training set characterization, as it requires too many training sessions. Diagonal-wise grouping of coefficients provides a middle ground between the two: it captures a fine-grained behavior along increasing frequencies, and is also feasible in terms of experimental time. Recent studies on task-specific image compression have shown that diagonal-wise pruning provides successful results \cite{sirin2024,sirin2025,xu_cvpr2020,xu_phdthesis2021}. 

\textbf{Task-specific Image Compression.} Furthermore, there has been a growing interest in task-specific image compression, where images are compressed for a specific computer vision task \cite{sirin2024,sirin2025,du_arxiv2022,fu_unpublished2016,lo_mmasia2019,santos_icip20,xu_cvpr2020,xu_phdthesis2021}. This requires learning in the compressed domain. Studies have shown that training and test images can be significantly compressed without significant loss of accuracy. This allows  reducing training storage, training time, and inference time, providing a resource-efficient computer vision system. Existing task-specific image compression studies are all based on a few standard benchmarks, similar to existing frequency characterization studies. Our characterization of the training and test sets sheds light on task-specific image compression across numerous application domains.

\textbf{Simplicity Bias.} Simplicity bias might occur due to visible \cite{beery2018,boland2024} or invisible superficial cues \cite{shortcut1,shortcut2}. While visible features are easy to detect via visual inspection, invisible features are deeply embedded in the data and require formal tools to analyze \cite{papernot2018, ramaswamy2023, bau2017, dabounou2024, zhao2019, shortcut1, shortcut2}. One such useful tool is Fourier analysis \cite{shortcut1,shortcut2}. Fourier analysis transforms data from its original domain, e.g., the spatial domain for images, into the frequency domain, enabling a principled analysis of shortcut behavior by leveraging the frequency structure embedded in the data. Analysis using other tools, such as geometric transformations or information-theoretic methods, is an interesting avenue for future work.

\begin{wrapfigure}{r}{0.45\linewidth}
  \centering
  \vspace{-0.8\baselineskip}
  \includegraphics[width=1\linewidth]{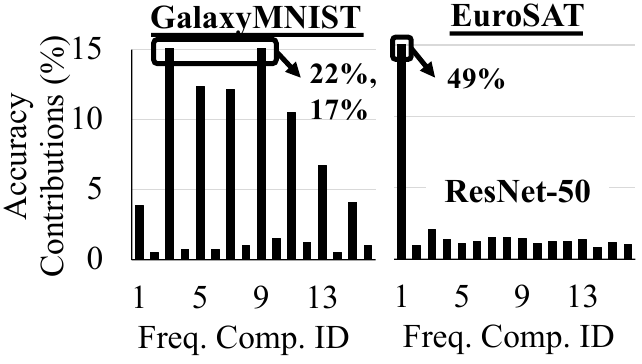}
  \caption{Skewed spectral behavior makes mixed-semantics tasks vulnerable to OOD corruptions.}
  \label{fig:mixed_sem_spec_behv}
  \vspace{-0.8\baselineskip}
\end{wrapfigure}

\section{Spectral Behavior of Mixed Texture-Shape Domains}
\label{sect:app:mixed_semantics}

Figure \ref{fig:mixed_sem_spec_behv} presents spectral behavior for GalaxyMNIST and EuroSAT datasets using ResNet-50. As can be seen, they are  both heavily skewed towards LFCs. GalaxyMNIST has a high contribution of every other LFCs with decreasing levels of contributions. EuroSAT has a dramatically high reliance to the lowest-frequency component. These skewed behaviors make GalaxyMNIST and EuroSAT vulnerable to OOD corruptions. Pruning LFCs allow network to learn from a wider-range of higher-frequency features and be more robust to OOD corruptions.

\section{Spectral Behavior of Pretrained Foundation Models}
\label{sect:app:vlm_vfm_spec_behv}

\begin{figure}[!h]
  \centering
   \includegraphics[width=1\linewidth]{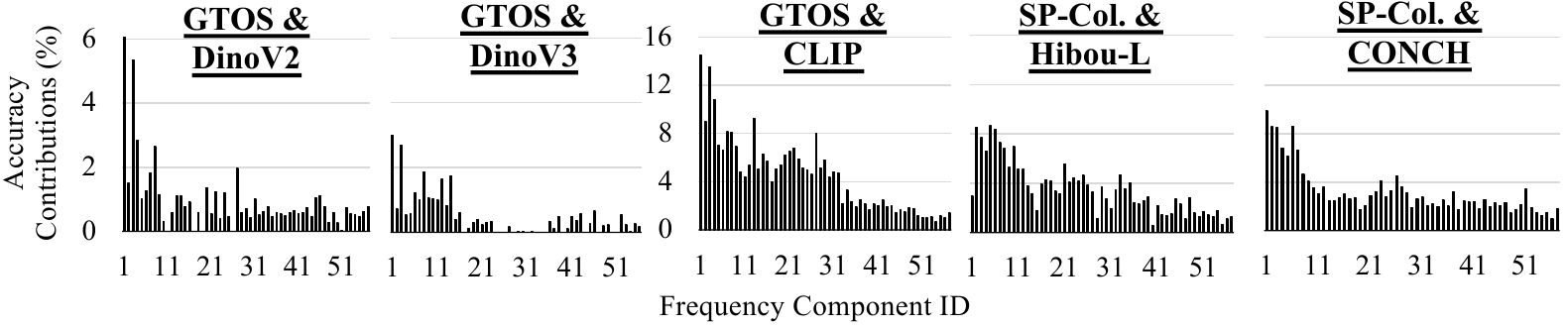}
   \caption{Spectral behavior of general-purpose (DinoV2, DinoV3, and CLIP) and domain-specific (Hibou-L and CONCH) pretrained foundation models.}
   \label{fig:spec_behv_vlm_vfm_all}
\end{figure}

Figure \ref{fig:spec_behv_vlm_vfm_all} presents spectral behavior of the five VLM/VFMs we analyze. As can be seen, all models have a skewed behavior towards LFCs, except DinoV3. Figure \ref{fig:all_vfms} has shown that ID and OOD accuracies of DinoV2 increases as we prune LFCs. Hence, DinoV2 (86M) suffers from low-frequency shortcuts, despite being trained with large number of images and complex data augmentations. DinoV3 (840M) mitigates the shortcuts thanks to its 10x larger model size compared to DinoV2. This, however, comes at the cost of a high computational cost. CLIP (151M) has a heavy reliance on the LFCs. As a result, pruning LFCs cause a significant accuracy losses, which it cannot recover using higher-frequency features. Hibou-L (300M) also has a high reliance on LFCs. It can, however, recover from the lost information when LFCs are pruned and maintain similar level of accuracy (Figure \ref{fig:all_vfms}). CONCH (307M), being the other VLM we analyze, is similar to CLIP. It heavily relies on LFCs, and it cannot recover from the lost information due to LFC-pruning.

CONCH’s unpruned accuracy (0.82) is lower than that of the from-scratch trained ResNet-50 (0.84). Once LFCs are pruned, ResNet-50 achieves a significantly higher accuracy than CONCH (0.88), despite its smaller size (25M) and smaller training set. This shows that, even though foundation models can be helpful for shortcuts, they might also introduce confounding features, resulting in suboptimal accuracy.

Hibou-L's unpruned accuracy is 0.9. Once we prune LFCs, its accuracy remains stable, then slowly drops. This shows that domain-specific foundation models can indeed mitigate the shortcuts. Nevertheless, (i) they may introduce confounding features in their pre-training set, as in CONCH, and (ii) they are very expensive to pre-train, fine-tune, and use for inference, with a few percent advantage compared to our from-scratch trained models. Our LFC-pruned ResNet-50 (25M params) achieves 0.88 accuracy, despite being >10x smaller and trained on a much smaller set.

\begin{wrapfigure}{r}{0.62\linewidth}
  \centering
  \vspace{-0.8\baselineskip}
  \includegraphics[width=1\linewidth]{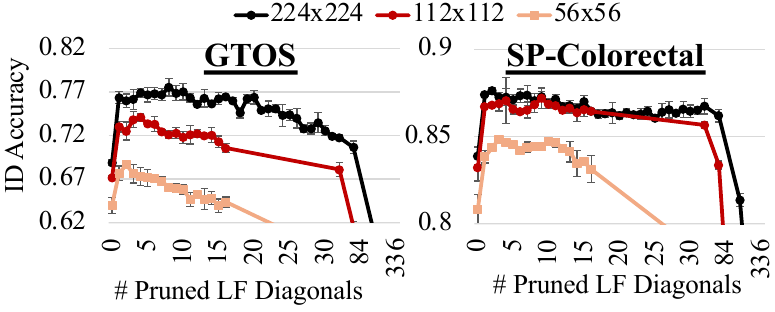}
  \caption{Low-frequency shortcuts persist across different image resolutions.}
  \label{fig:resolution_id}
  \vspace{-0.8\baselineskip}
\end{wrapfigure}

\begin{figure}[!h]
  \centering
   \includegraphics[width=.875\linewidth]{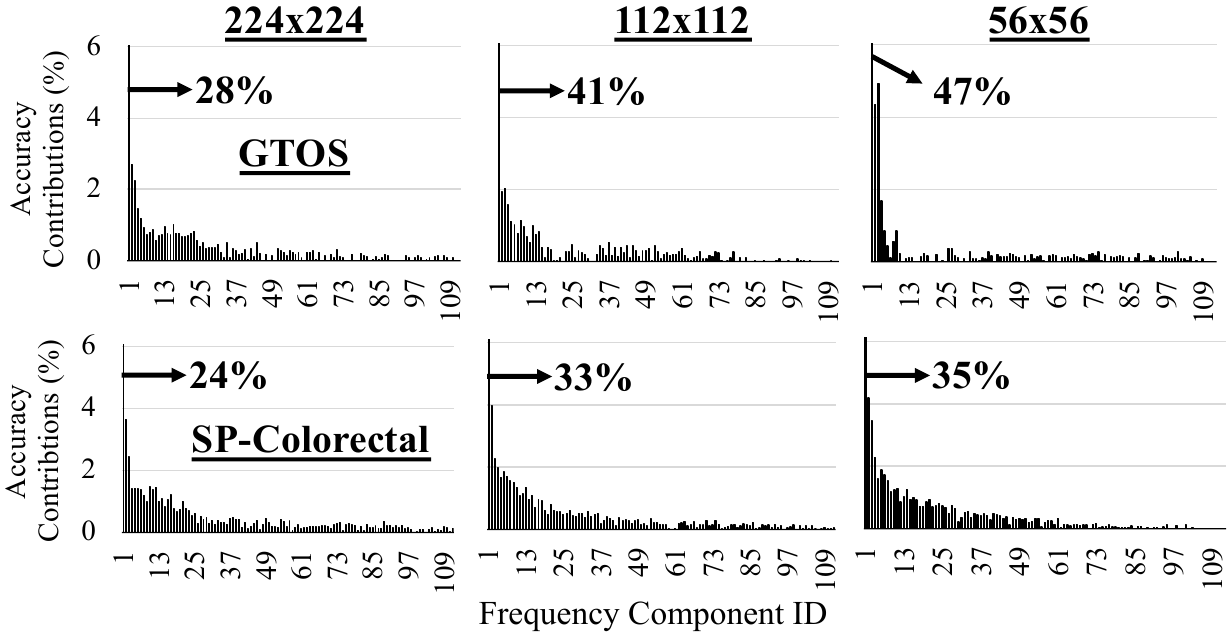}
   \caption{Lower-resolution images suffer from an increased the degree of low-frequency shortcuts, as they concentrate information on lower-frequency bands.}
   \label{fig:resolution_spec_behv}
\end{figure}

\section{Results on Image Resolution}
\label{sect:app:resolution}

Figure \ref{fig:resolution_id} presents ID accuracy results as we prune LFCs, for GTOS and SP-Colorectal using ResNet-50 and decreasing image resolutions from 224x224, to 112x112 and 56x56. As can be seen, pruning LFCs improves ID accuracy for both cases for all the image resolutions. Figure \ref{fig:resolution_spec_behv} presents the spectral behavior for unpruned trained for decreasing resolutions for GTOS (top row) and SP-Colorectal (bottom row). As can be seen, spectral behaviors get more and more skewed towards LFCs, as image resolutions decreases. This shows that decreased resolution increases the degree of low-frequency shortcuts.

Chapter 5.1.3.1 of \cite{signal_processing_book_2011} studies image halving in the frequency domain. Reducing the resolution of an image is equal to (i) transforming the image into the frequency domain, (ii) taking its top-left NxM square of coefficients, where N and M are reduced spatial dimensions, and (iii) applying inverse frequency transformation using the reduced NxM basis functions (see Figure 5.5 of \cite{signal_processing_book_2011}). Therefore, reducing image resolution is equivalent to concentrating information in lower-frequency bands, which results in an increased degree of low-frequency shortcuts.

\begin{figure}[!h]
  \centering
   \includegraphics[width=.95\linewidth]{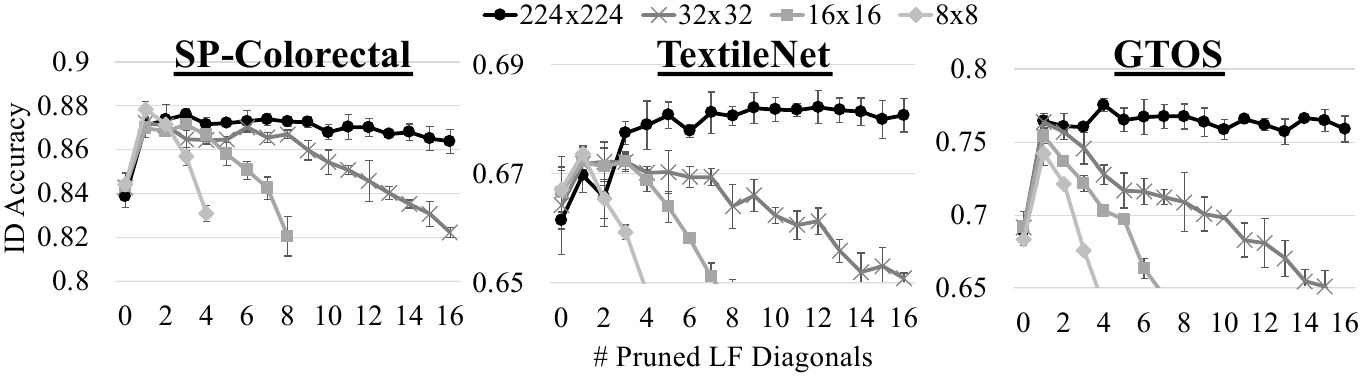}
   \caption{Large block sizes capture low-frequency shortcuts better than small block sizes.}
   \label{fig:block_size}
\end{figure}

\section{Results on Frequency-Transformation Block Size}
\label{sect:app:block_size}

Figure \ref{fig:block_size} presents ID accuracy results as we decrease block size from 224x224, to 32x32, 16x16, and 8x8. As can be seen, using larger block sizes generally produces higher accuracies than that of smaller block sizes. This is because low-frequency shortcuts are usually wide background cues that smaller block sizes only partially captures.

\begin{figure}[!h]
  \centering
   \includegraphics[width=.7\linewidth]{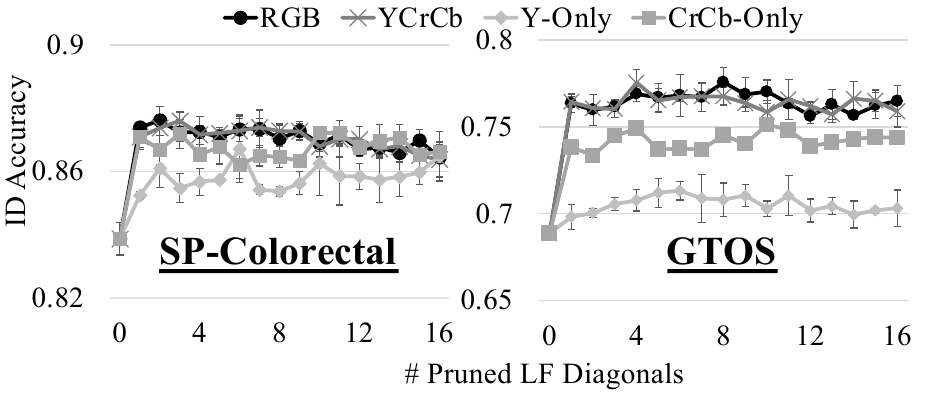}
   \caption{Low-frequency shortcuts persist across different color spaces. Pruning only grayscale or only color channels is not enough to completely eliminate the shortcuts.}
   \label{fig:color_spaces}
\end{figure}

\section{Results on Color Spaces}
\label{sect:app:color_spaces}

Figure \ref{fig:color_spaces} presents ID accuracy results when pruning LFCs over different color spaces and channels. RGB presents results for the color space that we use throughout the paper. YCrCb is a popular color space used by JPEG, which allows separating color information (Cr and Cr channels) from the grayscale information (Y channel). This is useful for many image compression, as human eye is more sensitive to the grayscale information that it is to the color information. Image compression algorithms more aggressively compress color channels than grayscale channel, providing a more efficient image storage \cite{wallace_tce92}.

As the figure shows, LFC-pruning over RGB or YCrCb color spaces have similar ID accuracy. On the other hand, pruning only the Y channel, or only the color channels have a lower accuracy than pruning all the YCrCb channels. This shows that low-frequency shortcuts is not a channel-specific phenomenon, but exits across all the channels. Hence, it needs to be mitigated across spectrum of all information channels.

\section{Frequency Characteristics of OOD Corruptions}
\label{sect:app:freq_char}

Figure \ref{fig:ood_freq_char} presents frequency characteristics of the used corruptions: fog, Gaussian blur, and elastic transform. We follow the same methodology as \cite{yin2019} when obtaining their Figure 2, with one exception. We use a log10 scale instead of using raw numbers. We use \texttt{torch.log10()}. This way, we can see fog's corruption at higher frequencies more easily, which would otherwise be invisible in Figure 2 of \cite{yin2019}. Other than this difference, we use the exact same methodology. Each image in Figure \ref{fig:ood_freq_char} shows the Fourier spectrum of the average difference between the original and corrupted images in the validation sets, for each dataset. The lowest-frequency coefficient is shifted to the center of each graph for compatibility with \cite{yin2019}. We use \texttt{torch.fft.fft2()}, and \texttt{torch.fft.fftshift()}.

\begin{figure*}[!ht]
  \centering
   \includegraphics[width=1\linewidth]{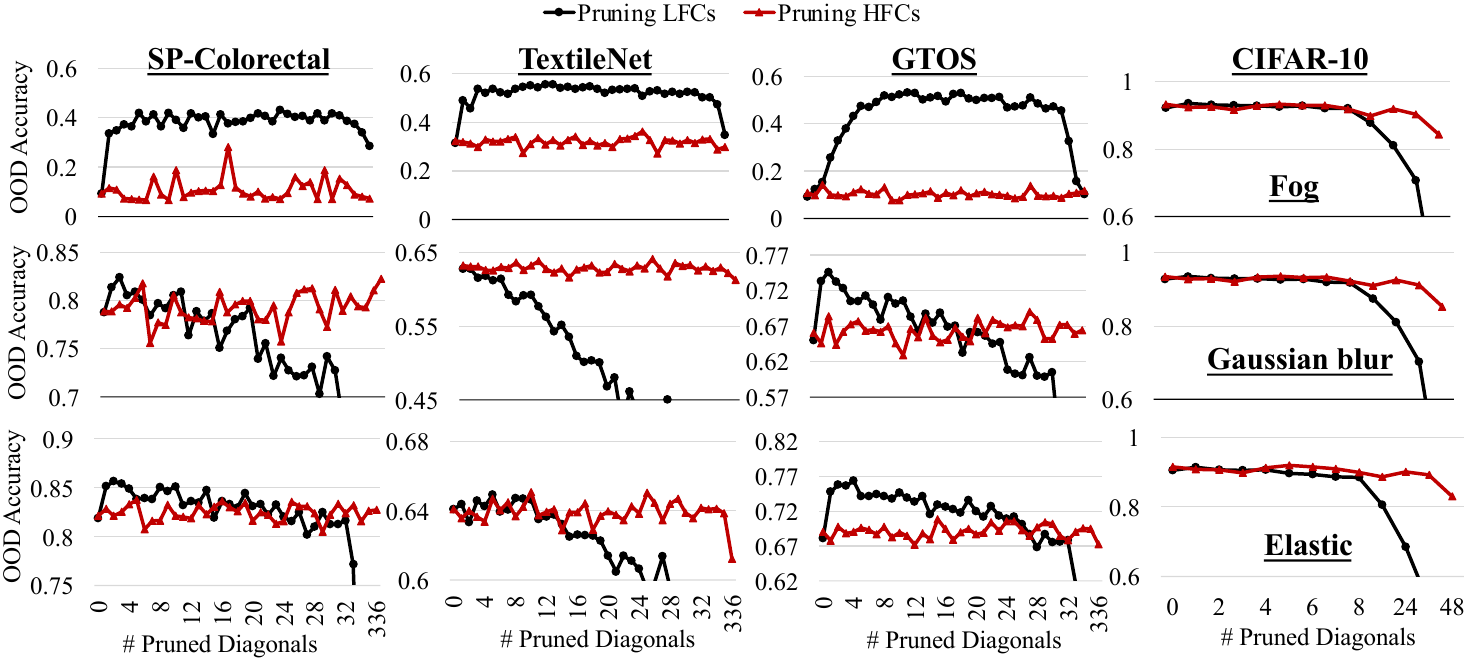}
   \caption{Impact of pruning HFCs on OOD performance (red line). We present pruning HFCs, along with the results for pruning LFCs, to provide a more complete picture. Similar to the ID performance, pruning HFCs has a minimal impact on OOD performance. For TextileNet and CIFAR-10, the impact is higher, as these two tasks depend more on higher frequencies at the lower end of the spectrum.}
   \label{fig:ood_hfc}
\end{figure*}

\section{Impact of HFCs on OOD Generalization}
\label{sect:app:hfc_ood}

Figure \ref{fig:ood_hfc} presents the impact of pruning LFCs (dark line) and HFCs (red line) on OOD generalization for the four tasks we analyze. As shown, pruning HFCs minimally impacts accuracy across all tasks and corruptions. This, once again, shows the efficiency of high-frequency compression. For CIFAR-10 and TextileNet, there is a slight drop at the end, similar to their ID accuracies. CIFAR-10 and TextileNet depend on higher frequencies, as their accuracy contribution graphs show in Figure \ref{fig:spec_behv_resnet50} (top row). Hence, pruning more than 224/32 diagonals from the bottom-right corner to the top-left corner reduces their accuracy.

%%%%%%%%%%%%%%%%%%%%%%%%%%%%%%%%%%%%%%%%%%%%%%%%%%%%%%%%%%%%

\clearpage

% \FloatBarrier

% \newpage

\bibliographystyle{plainnat}
\bibliography{main}

\end{document}